\documentclass{article}

\usepackage{arxiv}

\usepackage{makecell, tabularx}
\usepackage[export]{adjustbox}

\usepackage{algorithm}
\usepackage{tabularx}
\usepackage{rotating}
\usepackage[algo2e]{algorithm2e} 
\usepackage{algpseudocode}
\usepackage{url}
\usepackage[table,xcdraw]{xcolor}
\usepackage[colorlinks,linkcolor=blue,anchorcolor=blue,citecolor=green,urlcolor=black]{hyperref}
\usepackage[numbers]{natbib}
\usepackage{amsmath,amssymb}
\usepackage{bbm}
\usepackage{xcolor, soul}
\usepackage[utf8]{inputenc}
\usepackage[english]{babel}
\usepackage{graphicx}
\usepackage{booktabs}
\usepackage{multirow}
\usepackage{subcaption}
\usepackage{algpseudocode}
\usepackage{algorithm}
\usepackage{tabularray}
\usepackage[font=footnotesize]{caption}  
\newcolumntype{C}[1]{>{\centering\arraybackslash}p{#1}}
\renewcommand{\shorttitle}{ChangeMinds}

\usepackage{authblk}

\author[1,2]{Yuduo Wang}
\author[1,3]{Weikang Yu\thanks{Corresponding author}}
\author[4]{Michael Kopp}
\author[1,5]{Pedram Ghamisi}
\affil[1]{Helmholtz-Zentrum Dresden-Rossendorf (HZDR), 09599 Freiberg, Germany} 
\affil[2]{Humboldt-Universität zu Berlin, 10099 Berlin, Germany}
\affil[3]{Technical University of Munich, 80333 Munich, Germany}
\affil[4]{Nirmata Research}
\affil[5]{Lancaster University, LA1 4YR Lancaster, U.K.}

\date{}    
\title{ChangeMinds: Multi-task Framework for Detecting
and Describing Changes in Remote Sensing}

\begin{document}
\maketitle
\begin{abstract}
Recent advancements in Remote Sensing (RS) for Change Detection (CD) and Change Captioning (CC) have seen substantial success by adopting deep learning techniques. Despite these advances, existing methods often handle CD and CC tasks independently, leading to inefficiencies from the absence of synergistic processing. In this paper, we present \textit{ChangeMinds}, a novel unified multi-task framework that concurrently optimizes CD and CC processes within a single, end-to-end model. We propose the change-aware long short-term memory module (ChangeLSTM) to effectively capture complex spatiotemporal dynamics from extracted bi-temporal deep features, enabling the generation of universal change-aware representations that effectively serve both CC and CD tasks. Furthermore, we introduce a multi-task predictor with a cross-attention mechanism that enhances the interaction between image and text features, promoting efficient simultaneous learning and processing for both tasks. Extensive evaluations on the LEVIR-MCI dataset, alongside other standard benchmarks, show that ChangeMinds surpasses existing methods in multi-task learning settings and markedly improves performance in individual CD and CC tasks. Codes and pre-trained models will be available online at \href{https://github.com/Y-D-Wang/ChangeMinds}{https://github.com/Y-D-Wang/ChangeMinds}.
\end{abstract}


\section{Introduction}
\begin{figure}
    \centering
    \includegraphics[width=.5\linewidth]{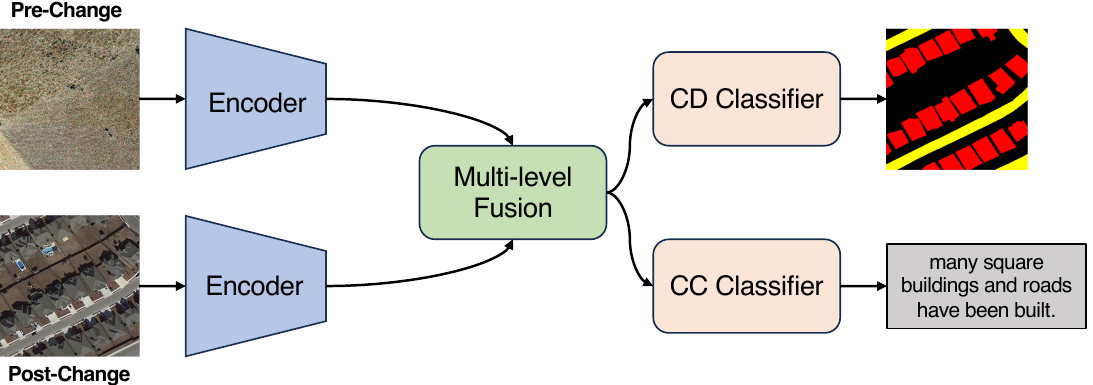}
    \caption{Multi-task learning framework overview for detecting and describing changes from bi-temporal RS images.}
    \label{fig:multi-task}
\end{figure}

%
%
%
%
Change detection (CD) has become a fundamental task in AI for Earth Observation (AI4EO) applications aimed at monitoring the dynamics of Earth's surfaces through remote sensing (RS) data \cite{8672156}. CD has contributed to a wide range of AI for social good initiatives \cite{ghamisi2024responsible,9690575}, including disaster monitoring \cite{zhang2023cross}, urban development, and environmental impact assessments \cite{yu2024maskcd}. Driven by recent advances in AI, CD approaches have evolved from traditional algorithms that rely on hand-crafted features to deep learning techniques capable of automatically predicting pixel-wise change maps from bi-temporal RS images using deep neural networks (DNNs) \cite{chang2024remote,hu2023binary,lei2020hierarchical,liu2017change}. Moreover, rapid developments in natural language processing (NLP) have led to the emergence of a new CD paradigm, change captioning (CC) \cite{tu2023adaptive}, which generates textual descriptions of change patterns in bi-temporal images, facilitating the interpretation of results in a readable format for RS experts. However, a clear gap exists between CD and CC tasks, as current approaches typically treat them as distinct tasks, developing models that address only one of them. Given that both CD and CC are critical for analyzing change patterns from different modalities, it is desirable to develop a more efficient strategy to tackle both tasks simultaneously within a unified architecture, as illustrated in Fig. \ref{fig:multi-task}.

Both CD and CC approaches share similar model designs, leveraging an encoder-decoder architecture that first extracts deep representations from bi-temporal images using a vision encoder, followed by text-based or pixel-based decoders for prediction. Specifically, the encoder often adopts a Siamese architecture involving a shared-weight vision backbone that extracts deep feature maps from bi-temporal images at multiple levels. The key difference lies in the decoders: the CD decoder retrieves high-level semantic information and reconstructs it into a change map, while the CC decoder interacts with the bi-temporal deep features, incorporating initialized text embeddings to generate change-aware captions. Since the distinction between CC and CD lies primarily in the decoder architecture, a unified framework that leverages multi-task learning is an appealing solution to bridge the gap between these tasks. Motivated by this, \cite{liu2024change} proposed a multi-task benchmark that includes the LEVIR-MCI dataset and a multi-stage model called MCINet, designed to facilitate both CC and CD tasks within a unified architecture. Although this approach offers a significant step toward unifying CC and CD with a common dataset, multi-task learning in this domain remains underdeveloped. MCINet, for example, fails to capture feature interactions between the CC and CD branches and does not offer an end-to-end solution, as it requires a three-stage training process that fine-tunes CC and CD tasks separately. Consequently, both tasks still face challenges in achieving optimal performance through complex, multi-step training processes.

To address these challenges, we propose a novel end-to-end multi-task learning method, \textit{ChangeMinds}, that bridges the gap between CC and CD, achieving optimal performance on both tasks simultaneously. Specifically, ChangeMinds consists of a Swin Transformer-based Siamese encoder, a ChangeLSTM module, and a multi-task predictor. The Siamese encoder extracts bi-temporal deep features from RS bi-temporal images, which are enhanced by the ChangeLSTM module to produce universal multi-level change-aware representations. The multi-task predictor first uses a unified change decoder to fuse these representations and then employs synergistic CC and CD branches to generate pixel-wise change maps and text-based change captions in parallel.

The main contributions of this paper are summarized as follows:

\begin{itemize} 
\item We present \textit{ChangeMinds}, an end-to-end multi-task framework that synergistically connects CD and CC tasks, allowing both tasks to contribute to each other through multi-task learning. 
\item We propose a ChangeLSTM module based on xLSTM architecture to capture complex spatiotemporal dynamics from two different directions, enabling the generation of universal change-aware representations that effectively serve both CC and CD tasks. 
\item We introduce a multi-task predictor that leverages these universal representations to effectively combine the strengths of CD and CC tasks through a cross-attention mechanism that facilitates rich interactions between textual and visual data. 
\item Experimental results on the LEVIR-MCI dataset and other common CC and CD datasets show that our proposed method achieves competitive performance in both multi-task learning and individual CC and CD tasks. 
\end{itemize}

The rest of this article is organized as follows: In Section \ref{s2}, a brief review is provided on deep learning-based CD methods, CC methods, and multi-task learning methods. Section \ref{s3} presents the details of the proposed ChangeMinds method. Section \ref{s4} covers details about the utilized
datasets and the experimental results, along with ablation studies. Finally, Section \ref{s5} concludes the paper.

\section{Related Work}\label{s2}

\subsection{Deep Learning-based RS Change Detection Methods}

\begin{figure}
    \centering
    \subfloat[\label{fig:CD}]{\includegraphics[width=.5\linewidth]{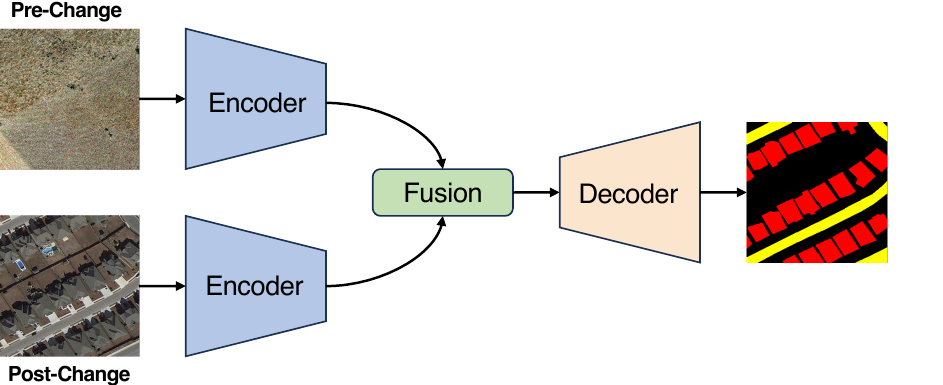}}
    \hspace{-2pt}
    \subfloat[\label{fig:CC}]{\includegraphics[width=.5\linewidth]{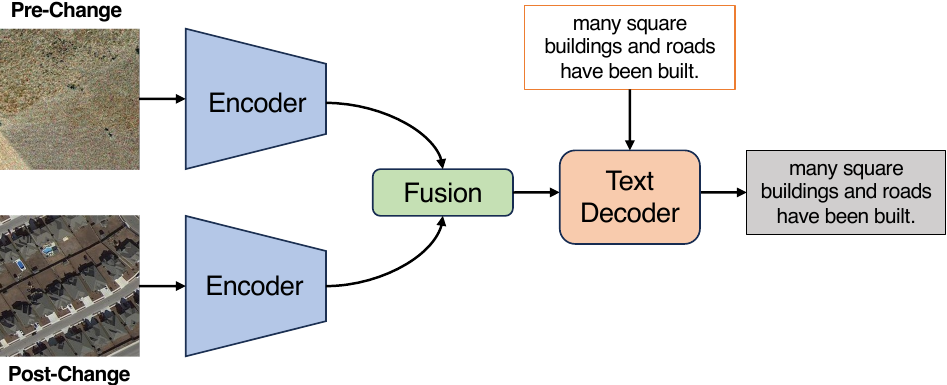}}
    \caption{Common Model Architectures for CD (a) and CC (b) tasks.}
\end{figure}

Deep learning techniques have significantly advanced CD methods by transitioning from hand-crafted features to deep feature extraction. This shift facilitates the automated generation of pixel-wise change maps from bi-temporal RS image pairs. As illustrated in Fig. \ref{fig:CD}, these approaches typically employ an end-to-end UNet-like architecture \cite{10616141}, featuring a Siamese encoder that extracts deep features from the bi-temporal images, followed by a decoder that reconstructs high-resolution change maps. For instance, \citet{daudt2018fully} introduced fully convolutional Siamese networks for CD, pioneering the application of deep neural networks (DNNs) in this domain. Their method extracts deep features using convolutional layers from the bi-temporal images, which are then differenced or concatenated to form change representations. These representations are subsequently reconstructed into change maps via transpose convolution and upsampling.

Building upon this architecture, numerous CD approaches have incorporated advanced modeling techniques, such as transformers and Mamba, to enhance the representation capabilities of DNNs and achieve more accurate change patterns. For example, \citet{zhang2022multilevel} proposed multi-level deformable attention-aggregated networks (MLDANets), which effectively learn long-range dependencies across multiple levels of bi-temporal deep features for multiscale context aggregation. \citet{10565926} introduced state space models for CD, developing a model named ChangeMamba that consists of a Mamba-based encoder and decoder. This model aims to fully learn global spatial contextual information from input images and facilitate spatiotemporal interactions of multitemporal features to obtain precise change information. Furthermore, \citet{yu2024minenetcd} proposed MineNetCD, a model based on a change-aware Fast Fourier Transform (ChangeFFT) module. This approach enables frequency domain learning to accurately capture change-aware representations from multi-level bi-temporal deep features.

Despite these advancements, existing approaches often suffer from inadequate context modeling capabilities and high computational complexity, leading to imbalanced performance when capturing complex spatiotemporal patterns embedded within bi-temporal deep features. Recent developments in extended long short-term memory (xLSTM) have demonstrated superior efficiency and performance in both context modeling and extrapolation compared to previous transformer techniques and state space models \citep{beck2024xlstm}. In this paper, we introduce a change-aware LSTM (ChangeLSTM) module designed to accurately model the intricate spatiotemporal patterns across bi-temporal deep features. By capturing long-range dependencies, ChangeLSTM obtains precise semantic information on change patterns, resulting in universal multi-level change-aware representations. These representations drive the accurate prediction of CD and CC tasks within the multi-task predictor of ChangeMinds.

\subsection{RS Change Captioning Methods}
In contrast to the CD, which aims to provide accurate change maps to indicate pixel-wise changes from bi-temporal RS images, CC aims to describe these changes in text form, making the information more accessible for human interpretation. Introduced by \citet{chouaf2021captioning} in 2021, CC has since attracted growing research attention, leading to the development of several datasets and methodologies. For instance, \citet{liu2022remote} constructed LEVIR-CC, a large-scale dataset derived from the well-known building CD dataset LEVIR-CD \cite{chen2020spatial}. Additionally, \citet{hoxha2022change} created an urban CC dataset based on the Dubai region to facilitate the understanding and automatic description of urbanization phenomena in that area.

Most CC methods adopt an encoder-decoder architecture similar to that used in CD tasks, as shown in Fig. \ref{fig:CC}. The encoder typically consists of a Siamese structure with shared weights to extract features from bi-temporal RS images, while the decoder transforms these features into textual descriptions through multimodal layers. To produce captions that effectively describe change patterns in text form, CC approaches generally initialize blank text embeddings and employ natural language processing (NLP) techniques such as recurrent neural networks (RNNs) \cite{chouaf2021captioning, hoxha2022change}, support vector machines (SVMs) \cite{hoxha2022change}, and transformers \cite{liu2022remote, chang2023changes}. These techniques facilitate the interaction between text embeddings and image features. For example, \citet{liu2022remote} developed RSICCformer, which utilizes a ResNet backbone to extract image features and incorporates a cross-attention mechanism alongside a multi-stage fusion module to progressively integrate these features. The fused features are subsequently passed to a transformer-based decoder to generate captions. Similarly, Chang and Ghamisi \cite{chang2023changes} proposed Chg2Cap, which leverages a hierarchical self-attention (HSA) block and ResBlocks to more effectively fuse the extracted feature maps. Furthermore, \citet{liu2023decoupling} introduced large language models (LLMs) into the RS-CC task and proposed a prompt-learning-based strategy for the efficient fine-tuning of LLMs. They also decoupled the RS-CC task into two sub-tasks, designing an image-level classifier and a discriminative feature encoder for each sub-task.

Although state-of-the-art approaches have led to notable performance improvements, most methods rely solely on features extracted from the final layer of the backbone, limiting the model's capacity to capture the full spatiotemporal semantic complexity of bi-temporal images. This often results in captions lacking detailed semantic information. To address this, our method leverages multi-level features extracted by the backbone. These features are refined by the ChangeLSTM module and fused within a unified change decoder as part of a multi-task predictor. This approach produces a universal, change-aware representation that captures rich contextual semantics, enhancing the accuracy of the generated captions.

\subsection{RS Multi-task Learning Methods}

While CD and CC have made considerable progress in recent years, these tasks are still typically handled independently during model development. A unified approach that integrates both tasks through multimodal processing of text and image features is highly desirable. Multimodality involves datasets containing not only images but also other data types, such as text, video, or audio. In the remote sensing (RS) community, multimodal tasks like RS Image Captioning \cite{zhao2021high}, RS Text-to-Image Generation \cite{xu2023txt2img} and RS Visual Question Answering \cite{wang2024rsadapter}, which combine image and text modalities, have attracted increasing attention. These tasks require models to understand both RS images and associated textual information, posing significant challenges. 

To address the need for a multimodal approach in CD and CC, \citet{liu2024change} introduced LEVIR-MCI, the first dataset designed for both tasks, which provides pixel-wise change maps and descriptive sentences for each pair of coregistered bi-temporal RS images. They also developed MCINet, a model with two independent branches: one for pixel-level (CD) and one for semantic-level (CC) change information. However, these branches function independently, with the CD branch utilizing multi-level features from the backbone, while the CC branch only uses deep features from the final layer. This lack of interaction between branches results in limited learning synergy and necessitates a multi-stage training process, as the CD and CC tasks are optimized separately. There are two main drawbacks in MCINet's architecture. First, the deep features extracted by the Siamese encoder are not universal across both tasks, requiring separate optimization for each, which prevents simultaneous learning. Second, the independent branches fail to interact during feature processing, leading to a loss of contextual information and undermining the potential for synergistic learning.

Our method overcomes these limitations by employing a multi-task predictor that integrates a unified change decoder and classifiers for CD and CC tasks. In this unified change decoder, multi-level change-aware representations are fused into a single, universal representation, which is used for both CD and CC predictions. This design not only enriches both classifiers with detailed semantic information but also uses a cross-attention layer to associate image features with text tokens. As a result, our model can be trained in an end-to-end, multi-task learning framework, enabling simultaneous learning and improving the overall performance of both tasks.

\section{Methodology}\label{s3}

\begin{figure*}[htb]
    \centering
    \includegraphics[width=\textwidth]{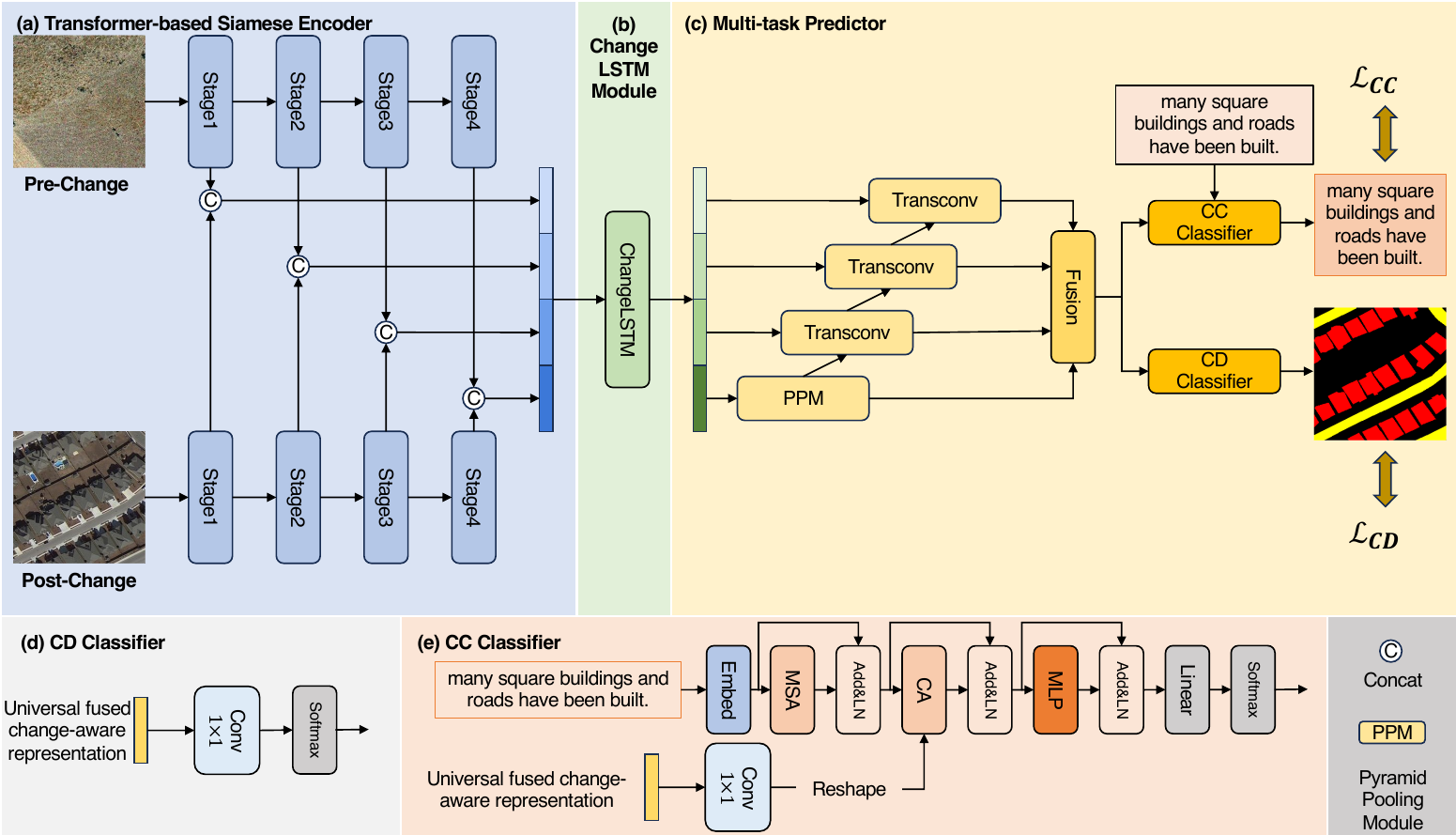}
    \caption{The overall structure of our proposed ChangeMinds for detecting and Describing Changes. (a) Transformer-based Siamese Encoder. (b) Change-aware LSTM (ChangeLSTM). (c) Multi-task Predictor. (d) CD Classifier and (e) CC Classifier are within the Multi-task Predictor.}
    \label{fig:flowchart}
\end{figure*}

This section provides a detailed description of our proposed unified end-to-end framework, ChangeMinds, as shown in Fig. \ref{fig:flowchart}. ChangeMinds consists of three main components: the Transformer-based Siamese Encoder, the ChangeLSTM module, and the Multi-task Predictor. After extracting and refining the bi-temporal deep features through the Siamese Encoder and ChangeLSTM, the universal change-aware representations are fed into the Unified Decoder within the Multi-task Predictor to generate a universal fused change-aware representation. This representation is then simultaneously passed into classifiers for CD and CC tasks to generate change maps and change captions concurrently.

\subsection{Transformer-based Siamese Encoder}
In recent years, transformer architectures have demonstrated excellent performance and are widely used as vision encoders in RS CD and CC tasks \cite{yu2024maskcd, liu2023progressive}. Similarly, we use a transformer-based backbone to build a Siamese encoder to extract multi-level deep features from bi-temporal images. 
In general, given a pair of bi-temporal images $(F^{1}, F^{2})$, where $F^1$ is the pre-change image and $F^2$ is the post-change image, the Siamese encoder, with weight-shared transformer-based backbones, progressively extracts multi-level deep features from both images, denoted as $\{F_{l}^{1}\}^{4}_{l=1}$ and $\{F_{l}^{2}\}^{4}_{l=1}$, respectively.
Subsequently, the extracted multi-level bi-temporal deep features are concatenated along the channel dimensions to form multi-level change features $\{F_{l}\}^{4}_{l=1}$ as follows:
\begin{equation}
    \{F_{l}\}^{4}_{l=1}=\{[F_{l}^{1};F_{l}^{2}]\}_{l=1}^{4}.
\end{equation}

In our model, we use the Swin Transformer \cite{liu2021swin} as the vision backbone to construct the Siamese encoder.
The Swin Transformer backbone consists of four stages, each containing multiple Swin Transformer blocks that progressively extract features. It employs a shifted-window self-attention mechanism, which reduces computational costs and preserves the locality of deep features by restricting global self-attention to sliding window regions.
The self-attention process in the shifted-window self-attention mechanism is calculated as follows:
\begin{equation}
	\text{Attention}(Q,K,V) = \text{SoftMax}\left(\frac{QK^{T}}{\sqrt{d}} + B\right)V,
\end{equation}
where $Q, K, V \in \mathbb{R}^{M^2 \times d}$ denote the query, key, and value matrices, $d$ is the dimension of the query and key, $M^2$ is the number of patches in a window and $B \in \mathbb{R}^{M^2\times M^2}$ is a trainable matrix which serves as the relative position bias. Each Swin Transformer block combines two types of self-attention: window-based multi-head self-attention (W-MHSA) and shifted window-based multi-head self-attention (SW-MHSA). W-MHSA captures context dependencies within local windows, while SW-MHSA enables the model to capture long-range dependencies across windows by applying position bias.
Following the Vision Transformer (ViT) \cite{dosovitskiy2020image}, each W-MHSA and SW-MHSA module is followed by a multi-layer perceptron (MLP) consisting of a linear layer expanding the input dimension by a factor 4, applies the GELU activation function and, finally, uses a second linear layer to project the resulting output into the original input dimension. The computation within one Swin Transformer block is expressed as:
\begin{align}
    X_{l}^{\prime} &= X_{l-1} + \text{W-MHSA}(\text{LN}(X_{l-1})), \\
    X_{l} &= X_{l}^{\prime} + \text{MLP}(\text{LN}(X_{l}^{\prime})), \\
    X_{l+1}^{\prime} &= X_{l} + \text{SW-MHSA}(\text{LN}(X_{l})), \\
    X_{l+1} &= X_{l+1}^{\prime} + \text{MLP}(\text{LN}(X_{l+1}^{\prime})).
\end{align}
where \(X_{l-1}\) and \(X_{l-1}^{\prime}\) denote the inputs to the W-MHSA module and MLP module of the \(l\)-th Swin Transformer block. \(X_{l}\), \(X_{l}^{\prime}\), and \(X_{l+1}^{\prime}\), \(X_{l+1}\) denote the inputs and outputs of the SW-MHSA and MLP modules in the \((l+1)\)-th Swin Transformer block. LN denotes the layer normalization layer. In each module, a residual connection is added after the operation to improve stability during training.

\subsection{ChangeLSTM Module}

\begin{figure*}
    \centering
    \includegraphics[width=.6\linewidth]{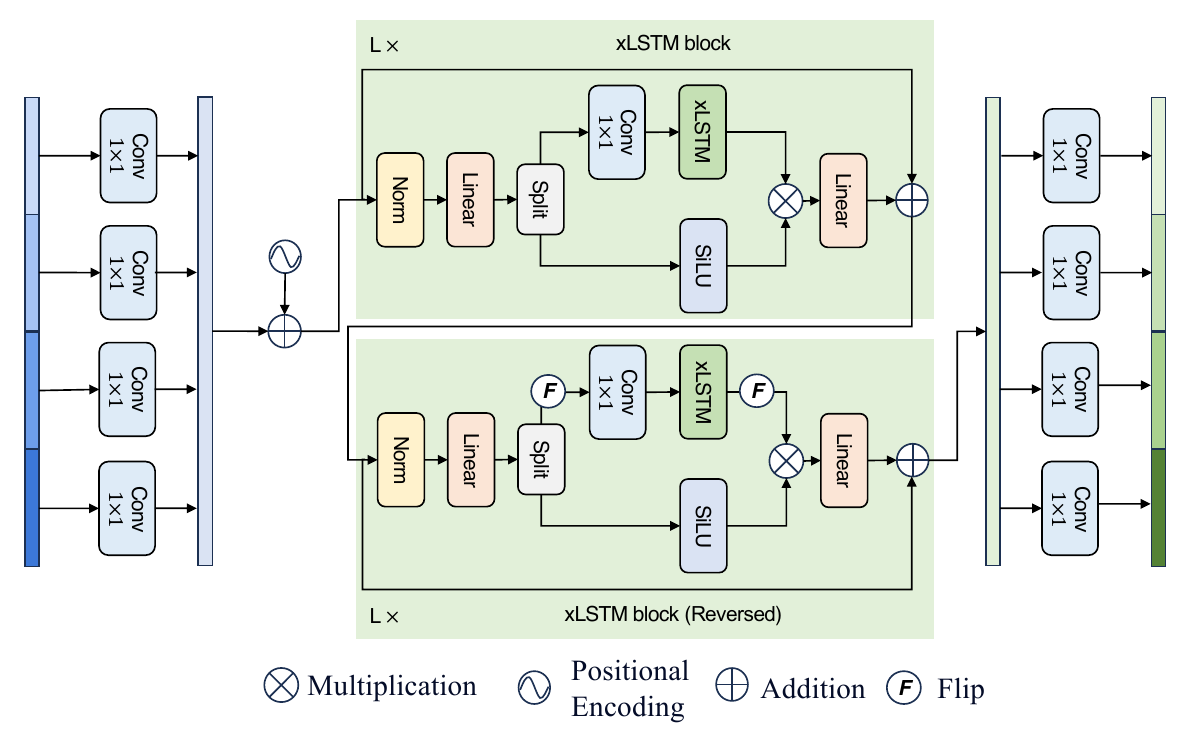}
    \caption{Illustration of the proposed ChangeLSTM module.}
    \label{fig:changelstm}
\end{figure*}

Recent advances in the xLSTM model, as introduced by \citet{beck2024xlstm}, have shown its effectiveness in handling large-scale data for NLP tasks through the incorporation of novel gating mechanisms and enhanced memory structures. Building on this, \citet{alkin2024vision} developed Vision-LSTM (ViL) based on the xLSTM block, successfully adapting the architecture for the image domain. Their work demonstrated the superiority of xLSTM models in various vision applications, particularly due to their ability to model long-range dependencies in deep features through the memory mechanism. In this paper, we leverage this strength to model the spatiotemporal dependencies in bi-temporal deep features, enabling more accurate learning of universal change-aware representations for subsequent CD and CC tasks.

In this paper, we introduce the xLSTM model to the remote sensing community for the first time by developing the ChangeLSTM module, which is based on the xLSTM block, as shown in Fig. \ref{fig:changelstm}. This module efficiently models the contextual information embedded in the multi-level change features extracted by the encoder. To enable interaction across these multi-level change features, we incorporate several \(1 \times 1\) convolutional layers to project the multi-level change features into a unified hidden dimension \(d_c\). The features are then flattened and concatenated to form a long-sequence deep feature \(\delta \in \mathbb{R}^{d_c \times \left(\sum_{l=1}^{4} h_l w_l\right)}\), which encapsulates multi-level semantic information. Here, \(h_l\) and \(w_l\) denote the height and width of the bi-temporal change features at the \(l\)-th level. To further preserve the spatial and hierarchical structure of these features, we add position and level embeddings to the corresponding deep feature points.

The xLSTM block consists of \(2L\) xLSTM blocks, where odd-numbered and even-numbered xLSTM blocks process tokens in opposite directions by flipping the deep feature sequence, where \(L\) denotes the depth of the ChangeLSTM module. This alternating directional processing enables the model to more effectively capture universal change-aware representations, thereby enhancing the model's performance in obtaining spatiotemporal patterns in CD and CC tasks. The processing in the xLSTM block with a forward direction can be expressed as follows:
\begin{align}
    [\delta_1; \delta_2] &= \text{LN}(\delta) \cdot W, \\
    \hat{\delta_1} &= \text{xLSTM}(\sigma(\text{Conv}(\delta_1))), \\
    \hat{\delta_2} &= \sigma(\delta_2), \\
    \hat{\delta} &= \hat{\delta_1} \odot \hat{\delta_2} + \delta.
\end{align}
where \(\text{LN}\) denotes the layer normalization, \(W\) represents a linear layer \(\mathbb{R}^{d_c \times 2d_c}\), and \(\text{Conv}(\cdot)\) denotes a \(1 \times 1\) causal convolutional layer. The function \(\sigma(\cdot)\) is the Sigmoid Linear Unit activation function (SiLU) \cite{elfwing2018sigmoid}, and xLSTM is the improved version of LSTM introduced in \cite{alkin2024vision}. 

After obtaining the change features refined by the ChangeLSTM module, we introduce several \(1 \times 1\) convolutional layers again to transform the features back to the original channel dimensions corresponding to each level. To this end, the multi-level change-aware representations \(\{\hat{\delta}_l\}_{l=1}^{4}\) are obtained through the ChangeLSTM module.

\subsection{Multi-task Predictor}

The multi-level change-aware representations \(\{\hat{\delta}_l\}_{l=1}^{4}\) are subsequently fed into the Multi-task Predictor, which comprises a unified change decoder based on UperNet \cite{xiao2018unified} and multi-task classifiers for both CD and change captioning CC tasks, as depicted in Fig. \ref{fig:flowchart}. The unified change decoder integrates a Pyramid Pooling Module (PPM) and a Feature Pyramid Network (FPN) to progressively reconstruct high-resolution semantic information from the multi-level change-aware representations, ultimately fusing them into universal change-aware representations \(y \in \mathbb{R}^{d \times H \times W}\), where \(H\) and \(W\) correspond to the original resolution of the bi-temporal images, \(256 \times 256\).

After that, we design two classifiers to predict results for the CD and CC tasks. The universal change-aware representations act as a bridge, feeding into both the CD and CC classifiers simultaneously. This design not only connects the two branches for joint output generation but also transfers multi-level semantic information, enabling the generation of more precise change captions.


Specifically, for the CD classifier, the universal change-aware representations are fed into a \(1 \times 1\) convolutional layer to predict change maps. The predicted change map probabilities \(P_{CD}\) can be expressed as follows:
\begin{equation}
    P_{CD} = \text{Softmax}(\text{Conv}(y)),
\end{equation}

For the CC Classifier, let \(T_t\) denote the \(t\)-th token of the input sentence, we first use an embedding layer to map the input text tokens \(T=[T_0, \dots, T_C]\) to a word embedding, followed by a multi-head self-attention (MSA) layer to extract global dependencies from the text tokens. The calculation process described above can be expressed as follows:
\begin{align}
    Z_{0} &= \text{Embed}(T) + P, \\
    Z_{t} &= \text{LN}(Z_{t-1} + \text{MSA}(Z_{t-1})),
\end{align}
where \(P\) is position embedding, \(Z_0\) is the initial input of MSA, \(Z_{t}\) and \(Z_{t-1}\) is the output and input of MSA  generating \(t\)-th token.

In existing CC methods, the interaction between image features and text tokens typically involves only high-level change features, which leads to the loss of detailed semantic information present in the multi-level deep features extracted by the backbone. To address this limitation, we employ a cross-attention layer to learn a simplicial linear combination of projected universal change-aware representations from a query of the relevant text tokens. Specifically, in the CC classifier, we first use a \(1\times 1\) convolutional layer to project \(y\) from the image domain to the text domain, facilitating a more robust fusion between the two modalities. Then, we reshape it to match the same hidden dimension of the text tokens and use it as the key ($K$) and value ($V$) in the multi-head cross-attention layer (CA) to fuse it with the text tokens. This fusion of image and text features allows for more effective interaction between the two modalities, resulting in the generation of more accurate change captions. The process can be described as follows:
\begin{align}
    y^{\prime} &= \text{Conv}(y),\\
    Z_{t}^{\prime} &= \text{LN}(Z_{t} + \text{CA}((Z_{t}), y^{\prime}, y^{\prime})),
\end{align}
where \(\text{Conv}(\cdot)\) denotes the \(1 \times 1\) convolutional projection layer, CA can be defined as:
\begin{align}
    \text{CA}((Z_{t}), y^{\prime}, y^{\prime})) &= \text{Concat(Head$_1$, $\dots$, Head$_h$)} W^O, \\
    \text{Head$_i$} &= \text{Softmax}(\frac{(Z_{t} W_{i}^{Q})(y^{\prime}W_{i}^{K^\mathrm{T}})}{\sqrt{d_k}}) (y^{\prime}W_{i}^{V}),
\end{align}
where $\text{Head$_i$}$, $W_{i}^{Q} \in \mathbb{R}^{d \times \frac{d}{h}}$, $W_{i}^{K} \in \mathbb{R}^{d \times \frac{d}{h}}$ and $W_{i}^{V} \in \mathbb{R}^{d \times \frac{d}{h}}$ are the $i$-th head cross attention and parameter matrices. $W^O \in \mathbb{R}^{d \times d}$ is the projection matrix. The feature \(Z_{t}^{\prime}\) is then passed through an MLP layer to further enhance the model's nonlinear representation capability:
\begin{equation}
    \hat{T_{t}} = \text{LN}(Z_{t}^{\prime} + \text{MLP}(Z_{t}^{\prime})),
\end{equation}

Finally, we feed \(\hat{T_{t}}\) into a linear layer followed by a softmax layer to obtain the final generated word probabilities \(P_{CC}\):
\begin{equation}
    P_{CC} = \text{Softmax}(\hat{T_{t}} \cdot w).
\end{equation}

\subsection{Multi-task Loss Function}

During training, the model generates predictions for both tasks concurrently, and we design a multi-task loss function to ensure balanced learning across both objectives. This approach allows the model to leverage shared features while maintaining task-specific performance.

\subsubsection{CD Loss}

For the CD task, we use the widely adopted cross-entropy loss to measure the discrepancy between the predicted change map probabilities and the ground truth. The loss for the CD task is defined as:
\begin{equation}
\mathcal{L}_{\text{CD}} = - \sum_{h=1}^{H} \sum_{w=1}^{W} \sum_{c=1}^{C} y_{c}^{hw} \log(p_{c}^{hw}),
\end{equation}
where \(y_{c}^{hw}\) and \(p_{c}^{hw}\) are the ground truth label and the predicted change probability for position \((h,w)\), respectively. Here, \(C\) denotes the number of classes, while \(H\) and \(W\) are the height and width of the ground truth.

\subsubsection{CC Loss}

For the CC task, we similarly employ cross-entropy loss to measure the discrepancies between the predicted change caption probabilities and the ground truth captions. The loss for the CC task is defined as:
\begin{equation}
\mathcal{L}_{\text{CC}} = -\sum_{m=1}^{M} \sum_{n=1}^{N} y_{m,n} \log(p_{m,n}),
\end{equation}
where \(y_{m,n}\) and \(p_{m,n}\) represent the \(m\)-th word in the corresponding descriptive sentences and the predicted probability of the \(m\)-th word at time step \(n\), respectively. Here, \(M\) is the length of the predicted change caption, and \(N\) is the size of the vocabulary.

\subsubsection{multi-task Loss}

As indicated in \cite{he2022metabalance}, a larger gradient magnitude signifies a greater impact on model updates. To effectively balance the losses between the two tasks and ensure that both contribute equally to model optimization, we employ the method described in \cite{he2022metabalance} for multi-task learning. The combined loss can be expressed as:
\begin{equation}
    \mathcal{L} =  \mathcal{L}_{\text{CC}} + \mathcal{L}_{\text{CD}} \cdot \frac{\text{detach}(\mathcal{L}_{\text{CC}})}{\text{detach}(\mathcal{L}_{\text{CD}})}.
\label{loss}
\end{equation}
where \(\text{detach}(\cdot)\) denotes the detach operation. This approach ensures that the contributions of each task's loss to the parameter updates are effectively balanced.

\section{Experimental Results}\label{s4}

\subsection{Dataset Descriptions}

In this work, we evaluate the performance of ChangeMinds under multi-task learning settings on the LEVIR-MCI dataset. Additionally, we assess the method's effectiveness on five common CD datasets and the DUBAI-CC dataset, evaluating its performance on both the CD and CC tasks separately. The images in all datasets are resized to \(256 \times 256\) pixels to ensure a consistent and fair comparison.

\subsubsection{LEVIR-MCI Dataset}

The LEVIR-MCI dataset \cite{liu2024change} comprises 10,077 bi-temporal images, with a resolution of \(256 \times 256\) pixels, covering over 20 regions in Texas, USA. Each image pair is annotated by corresponding CD masks, which include roads and buildings, and five descriptive sentences. To ensure a fair comparison of model performance, we split the dataset as described in \cite{liu2024change}, utilizing 6,815 bi-temporal images as the training set, 1,333 bi-temporal images as the validation set, and the remaining 1,929 bi-temporal images as the testing set.

\subsubsection{CD Datasets}

To evaluate the model's performance on the individual CD task, we use five common CD datasets. The characteristics of each dataset are as follows:
\begin{itemize}
    \item \textbf{CLCD \cite{liu2022cnn}} comprises 600 pairs of bi-temporal agricultural images, each with a resolution of \(512 \times 512\) pixels and spatial resolutions varying between 0.5 and 2 meters.
    \item \textbf{EGY-BCD \cite{holail2023afde}} contains 6,091 bi-temporal image pairs, each with a resolution of \(256 \times 256\) pixels.
    \item \textbf{GVLM-CD \cite{zhang2023cross}} consists of 7,496 pairs of patches, each measuring \(256 \times 256\) pixels.
    \item \textbf{LEVIR-CD \cite{chen2020spatial}} includes 637 bi-temporal image pairs, also with a size of \(256 \times 256\) pixels.
    \item \textbf{SYSU-CD \cite{shi2021deeply}} consists of 20,000 bi-temporal image pairs, each with a size of \(256 \times 256\) pixels.
\end{itemize}

\subsubsection{CC Dataset}

To evaluate the proposed method specifically on the CC task, we use the DUBAI-CC dataset \cite{hoxha2022change}. This dataset contains 500 bi-temporal image pairs, each with a resolution of \(50 \times 50\) pixels. Each bi-temporal image pair is accompanied by five manually annotated change captions. We follow the dataset split in the original paper that utilizes 300 samples as the training set, 50 samples as the validation set, and the remaining 150 samples as the testing set.

\subsection{Baseline Approaches}

In the multi-task learning settings, we compare the proposed method, ChangeMinds, with the current state-of-the-art method, MCINet \cite{liu2024change}. Since MCINet employs a multi-stage approach to enhance model performance, we reproduce its results under the same experimental setup for a fair comparison. Due to the lack of other methods specifically designed for multi-task learning, we also compare ChangeMinds with seven additional state-of-the-art methods focused on the CD task, including BiT \cite{chen2021remote}, ChangeFormer \cite{bandara2022transformer}, DMINet \cite{feng2023change}, ICIFNet \cite{feng2022icif}, RDPNet \cite{chen2022rdp}, Siam-Diff \cite{daudt2018fully}, and SNUNet \cite{fang2021snunet}. Furthermore, we assess the performance of five state-of-the-art methods targeting the CC task: DUDA \cite{park2019robust}, MCCFormer-S and MCCFormer-D \cite{qiu2021describing}, RSICCFormer \cite{liu2022remote}, and Chg2Cap \cite{chang2023changes}.

\subsection{Implementation Details}

In our experiments, we utilize Swin-T \cite{liu2021swin} as the Transformer-based Siamese encoder to extract features from the bi-temporal images. The unified hidden dimension \(d_c\) in the ChangeLSTM module is set to 256, and the depth of the xLSTM block \(L\) is set to 2. The universal fused change-aware representations hidden dimension \(d\) is set to 512. The maximum number of epochs is set to 50. We employ the Adam optimizer \cite{kingma2014adam}, with an initial learning rate of \(1 \times 10^{-4}\). To enhance the convergence, we utilize a cosine annealing scheduler that gradually reduces the learning rate to \(1 \times 10^{-7}\).

To ensure the reproducibility of the experiments, we set the random seed to 42. The batch size is configured to 8 per GPU, resulting in a total batch size of 32 across all GPUs. All experiments are conducted on the Slurm high-performance computing (HPC) system, which features a 128-core CPU and 4 NVIDIA Tesla A100 GPUs (80 GB of RAM). Additionally, we adopt the Accelerate \cite{accelerate} package to expedite the computational processes of the models in our multi-GPU environment.

\subsection{Evaluation Metrics}

Since the detection masks in the LEVIR-MCI dataset contain two different classes, we use the mean Intersection over Union (mIoU) metric for the CD task to keep consistent with MCINet \cite{liu2024change}. In the subsequent ablation experiments, where the CD datasets are binary classified, we use the common F1 score and change-class Intersection over Union (cIoU) as metrics to evaluate model performance. We employ commonly used metrics such as BLEU-n ($n=1,2,3,4$) \cite{papineni2002bleu}, METEOR \cite{banerjee2005meteor}, ROUGE-L \cite{lin2004rouge} and CIDEr-D \cite{vedantam2015cider} to evaluate the CC task. 

For the CD task in the multi-task learning setting, mIoU is calculated according to the following equations:

\begin{align}
    \mathrm{IoU}_{i} &= \frac{\mathrm{TP}_i}{\mathrm{TP}_i + \mathrm{FP}_i + \mathrm{FN}_i} \\
    \mathrm{mIoU} &= \frac{1}{\mathrm{N}} \sum_{i=1}^{\mathrm{N}} \mathrm{IoU}_{i} 
\end{align}
where N is the number of classes, the TP$_{i}$, FP$_{i}$, and FN$_{i}$ are the number of true positive, false positive, and false negative for class $i$, respectively. IoU$_{i}$ denotes the intersection over Union for class $i$.

For the CD task in ablation experiments, the F1 score and cIoU are calculated as:

\begin{align}
    \mathrm{F1} &= \frac{2\mathrm{TP}}{\mathrm{2TP+FP+FN}} \\
    \mathrm{cIoU} &= \frac{\mathrm{TP}}{\mathrm{TP} + \mathrm{FP} + \mathrm{FN}}
\end{align}
where the TP, FP, FN are the number of true positive, false positive, and false negative, respectively.

\subsection{Comparison with the state-of-the-art methods}

\begin{table*}
\caption{The comprehensive performance comparisons on the LEVIR-MCI dataset. The best results are highlighted in bold.}
\label{tab:Comparisons_other_methods}
\centering
\resizebox{.9\linewidth}{!}{
\begin{tabular}{c|l|c|c c c c c c c}
	\toprule
 \multicolumn{2}{c|}{\shortstack{Method}} & mIoU & BLEU-1 & BLEU-2 & BLEU-3 & BLEU-4 & METEOR & ROUGE-L & CIDEr-D \\
	\midrule
    \multirow{7}{*}{\shortstack{Change \\Detection}}
    & BiT \cite{chen2021remote} & 0.8314 & -- & -- & -- & -- & -- & -- & -- \\
    & ChangeFormer \cite{bandara2022transformer}& 0.7850 & -- & -- & -- & -- & -- & -- & -- \\ 
    & DMINet \cite{feng2023change} & 0.7927 & -- & -- & -- & -- & -- & -- & -- \\
    & ICIFNet \cite{feng2022icif} & 0.7926 & -- & -- & -- & -- & -- & -- & -- \\
    & RDPNet \cite{chen2022rdp} & 0.7461 & -- & -- & -- & -- & -- & -- & -- \\
    & Siam-Diff \cite{daudt2018fully} & 0.8375 & -- & -- & -- & -- & -- & -- & -- \\
    & SNUNet \cite{fang2021snunet} & 0.8276 & -- & -- & -- & -- & -- & -- & -- \\

    \midrule
    \multirow{5}{*}{\shortstack{Change \\Captioning}}
    &DUDA \cite{park2019robust} & -- &0.8144 &0.7222 &0.6424 &0.5779 &0.3715 &0.7104 &1.2432 \\
    &MCCFormer-S \cite{qiu2021describing} & -- &0.7990 &0.7026 &0.6268 &0.5668 &0.3617 &0.6946 &1.2039 \\
    &MCCFormer-D \cite{qiu2021describing} & -- & 0.8042 &0.7087 &0.6286 &0.5638 &0.3729 &0.7032 &1.2444 \\
    &RSICCFormer \cite{liu2022remote} & -- &    0.8472 & 0.7627 & 0.6887 & 0.6277 & 0.3961 & 0.7412 & 1.3412  \\
    &Chg2Cap \cite{chang2023changes} & -- &   0.8614 & 0.7808 & 0.7066 & 0.6439 & 0.4003 & 0.7512 & 1.3661  \\  
        \midrule
    \multirow{2}{*}{\shortstack{multi-task}}
    & MCINet \cite{liu2024change} & 0.8620 & 0.8584 & 0.7767 & 0.7060 & 0.6497 & 0.4010 & 0.7510 & 1.3776 \\
    & Ours & \textbf{0.8678} & \textbf{0.8639}  & \textbf{0.7834} & \textbf{0.7135} & \textbf{0.6560} & \textbf{0.4086} & \textbf{0.7585} & \textbf{1.4032}\\

	\bottomrule
\end{tabular}
}
\end{table*}

\subsubsection{Quantitative Comparisons}

Table \ref{tab:Comparisons_other_methods} presents the performance comparison between ChangeMinds and competitive methods on the LEVIR-MCI dataset. Since some methods in their original papers only addressed a single CD or CC task, we only compared the relevant models' performance on the corresponding task. The table shows that ChangeMinds outperforms other methods across all metrics. Specifically, compared to MCINet, ChangeMinds achieves a 0.58\% improvement on mIoU, a 0.63\% improvement on BLEU-4, and a 2.56\% improvement on CIDEr-D. Compared to the other seven CD methods, ChangeMinds achieves significant improvements of 3.03\%-12.17\% on mIoU. Moreover, ChangeMinds improves BLEU-4 by 1.21\% and CIDEr-D by 3.71\% compared to Chg2Cap. All these quantitative results indicate that using multi-level fused image features can effectively improve change detection and change captioning quality.

Additionally, the performance of ChangeMinds and MCINet significantly surpasses other single-task methods, which further demonstrates the superiority of the multi-task learning strategy.

\begin{figure*}
    \centering
    \includegraphics[width=.7\linewidth]{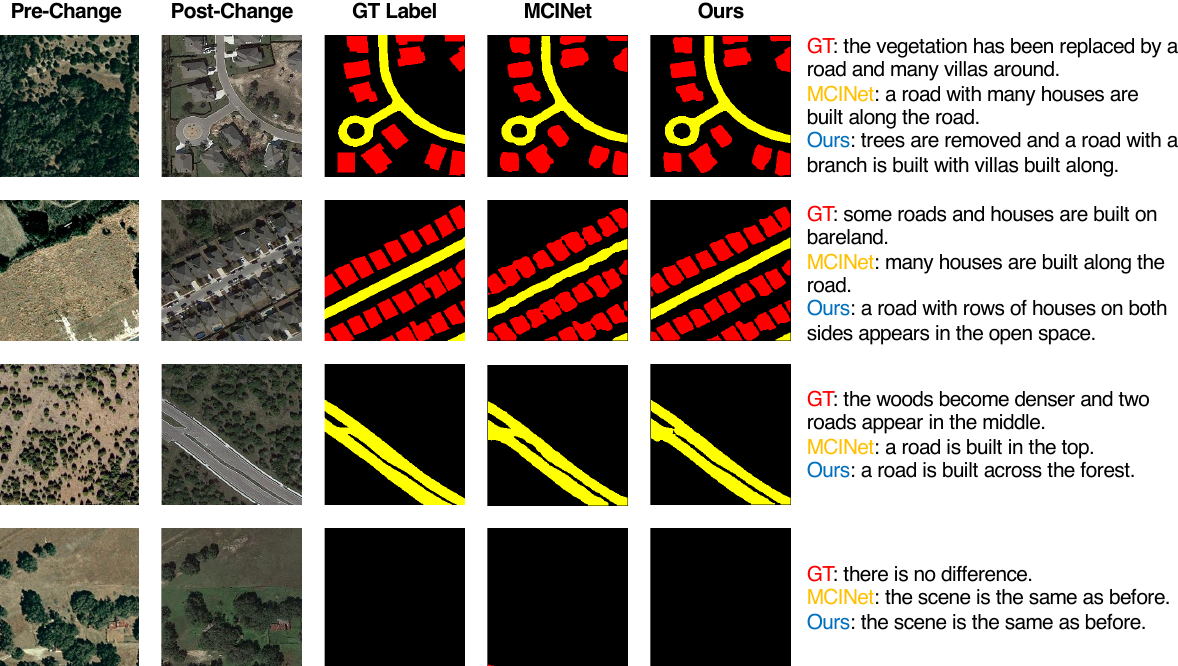}
    \caption{Qualitative comparisons of ChangeMinds and MCINet \cite{liu2024change} on the LEVIR-MCI dataset. The images from left to right are as follows: the pre-change image, the post-change image, the CD ground truth label, the change maps predicted by MCINet, the change maps predicted by ChangeMinds, and a comparison of the captions generated by MCINet and our method, with GT representing the annotated ground truth captions.}
    \label{MCI_comp}
\end{figure*}

\begin{figure}
    \centering    \includegraphics[width=.65\linewidth]{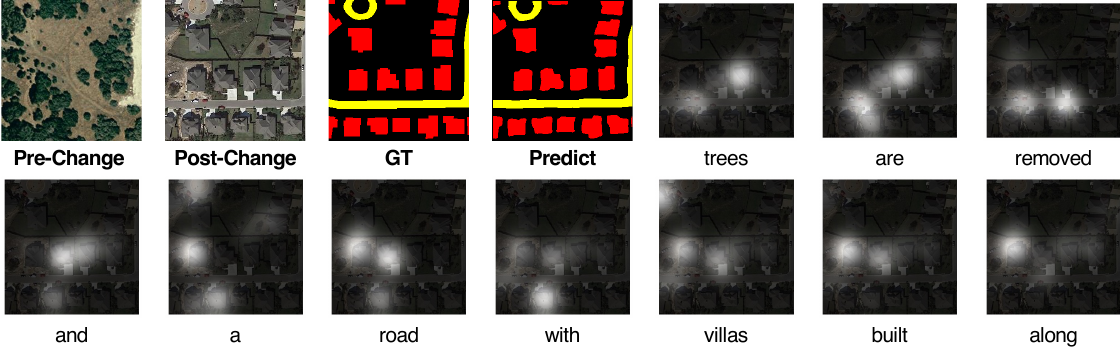}
    \caption{The visualization of attended images along with the caption generation processes of ChangeMinds on the LEVIR-MCI dataset.} 
\label{Caption_gen}
\end{figure}

\subsubsection{Qualitative Comparisons}

To evaluate the accuracy of predicted change maps and the quality of generated change captions, we show qualitative comparisons in Fig. \ref{MCI_comp}. Our method demonstrates superior capability in identifying small targets and subtle changes within the bi-temporal images. Specifically, in the first and second rows of Fig. \ref{MCI_comp}, the roads identified by our method are notably smoother, while those detected by MCINet are coarser. In the example from the third row, the change map generated by ChangeMinds provides a more accurate road width.

Additionally, our method can generate more detailed change captions and effectively identify the background of bi-temporal images. For instance, in the second row of Fig. \ref{MCI_comp}, ChangeMinds identifies that the houses are built at both ends of the road and the area was previously an open space. Similarly, in the next example, our method successfully recognizes that the background of the bi-temporal images is the forest. 

Furthermore, to better qualitatively evaluate the captions generated by our method, we visualize the image attention weights corresponding to each word in Fig. \ref{Caption_gen}. We can observe that our method can successfully capture specific regions of targets and generate correct descriptions.


\subsection{Ablation Studies}

\subsubsection{Effectiveness of multi-task learning}

\begin{table*}
\centering
\caption{Quantitative comparison between multi-task learning and single-task. The best results are highlighted in bold.}
\label{tab:multi-task}
\resizebox{.8\linewidth}{!}{
\begin{tabular}{l|c|ccccccc}
	\toprule
	Method & mIoU & BLEU-1 & BLEU-2 & BLEU-3 & BLEU-4 & METEOR & ROUGE-L & CIDEr-D \\
	\midrule
	{Only CD} & 0.8648  & -- & -- & -- & -- & -- & -- & -- \\
	{Only CC} & -- & 0.8535	&0.7748	&0.7061	& 0.6489	&0.3992 &0.7525	&1.3779	 \\
 
	{Multi-task Training} & \textbf{0.8678} & \textbf{0.8639}  & \textbf{0.7834} & \textbf{0.7135} & \textbf{0.6560} & \textbf{0.4086} & \textbf{0.7585} & \textbf{1.4032} \\
	\bottomrule
\end{tabular}}
\end{table*}

We verify the effectiveness of multi-task learning. Table \ref{tab:multi-task} gives the experimental results of single-task learning and multi-task learning. 
In this table, "only CC" and "only CD" indicate that the model is optimized only using \(\mathcal{L}_{\text{CD}}\) or \(\mathcal{L}_{\text{CC}}\). "Multi-task learning" refers to optimizing the model by using multi-task loss \(\mathcal{L}\).
We can observe that compared to optimizing the model with a single CC or CD loss, the multi-task learning approach achieves better results across all evaluation metrics. This experimental result proves that the introduction of multi-task learning can effectively enhance the interaction between the two branches, improving the model's ability to detect and describe changes.

\subsubsection{Effectiveness of ChangeLSTM}

To validate the effectiveness of the proposed ChangeLSTM module, we conduct several experiments, and list results in Table \ref{tab:changeLSTM}. "Model optimization only for CD task" and "Model optimization only for CC task" mean we only use \(\mathcal{L}_{\text{CD}}\) or \(\mathcal{L}_{\text{CC}}\) to optimize the model. "Multi-task learning" represents applying multi-task loss \(\mathcal{L}\). Comparing the three different settings, an improvement in CC and CD evaluation metrics can be observed when the model uses ChangeLSTM, indicating that modeling the intricate spatiotemporal patterns and capturing long-range dependencies are crucial to the model's reasoning ability.

Additionally, compared to the model using only \(\mathcal{L}_{\text{CD}}\) and \(\mathcal{L}_{\text{CC}}\), when the model uses ChangeLSTM with \(L=2\) and applies the multi-task loss \(\mathcal{L}\), it can simultaneously improve the performance of both CD and CC tasks. This demonstrates that the two classifiers in the model can contribute to each other and jointly enhance the model's overall performance. However, although the BLEU-4 and CIDEr-D scores are better when \(L=1\) compared to \(L=2\), the introduction of the multi-task loss \(\mathcal{L}\) leads to a decrease in mIoU score. Therefore, we set the \(L\) of the ChangeLSTM module to \(2\).

\begin{table}[htp]
\centering
\caption{Ablation study of ChangeLSTM on the LEVIR-MCI dataset.}
\label{tab:changeLSTM}
\resizebox{.5\linewidth}{!}{
\begin{tabular}{l|c|cccc}
	\toprule
	Variants & mIoU & B1 & B4  & R & C \\ \midrule
        \multicolumn{6}{l}{Model optimization only for CD task} \\ 
        \midrule
        w/o ChangeLSTM & 0.8634  & --  & --  & -- & -- \\
        w/ ChangeLSTM, L=1 & 0.8663  & --  & --  & -- & -- \\
        w/ ChangeLSTM, L=2 & 0.8648  & --  & --  & -- & -- \\
	\midrule
        \multicolumn{6}{l}{Model optimization only for CC task} \\ 
        \midrule
	w/o ChangeLSTM  & --  & 0.8600	& 0.6420 &0.7506	&1.3770 \\
	w/ ChangeLSTM, L=1 & -- & 0.8650	& 0.6536 &0.7577	&1.4127	 \\
        w/ ChangeLSTM, L=2 & -- & 0.8535	& 0.6489 &0.7525	&1.3779	 \\
        \midrule
        \multicolumn{6}{l}{Multi-task learning} \\
        \midrule
        w/o ChangeLSTM  & 0.8607  & 0.8615	& 0.6492 &0.7553	&1.3788 \\
	w/ ChangeLSTM, L=1 & 0.8651 & 0.8636	& 0.6583 &0.7622	&1.4166	 \\
        w/ ChangeLSTM, L=2 & 0.8678 & 0.8639	& 0.6560 &0.7585	&1.4032	 \\
	\bottomrule
\end{tabular}}
\end{table}

\subsubsection{Comparison with other CD methods}

\begin{table*}
\centering
\caption{Quantitative Comparison Results in terms of F1 and cIoU on five different CD Datasets. The best results are highlighted in bold.}
\resizebox{.8\linewidth}{!}{
\begin{tabular}{l|cc|cc|cc|cc|cc}
\toprule 
\multirow{2}{*}{Method} & \multicolumn{2}{c|}{\textbf{CLCD}} & \multicolumn{2}{c|}{\textbf{EGY-BCD}} & \multicolumn{2}{c|}{\textbf{GVLM}} & \multicolumn{2}{c|}{\textbf{LEVIR-CD}} & \multicolumn{2}{c}{\textbf{SYSU-CD}} \\ 
& F1 & cIoU  & F1 & cIoU & F1 & cIoU & F1 & cIoU & F1 & cIoU   \\ \midrule
BiT \cite{chen2021remote} & 0.6591 & 0.4915 & 0.7906 & 0.6538 & 0.8768 & 0.7806 & 0.8884 & 0.7992 & 0.7497 & 0.5997 \\ 
ChangeFormer \cite{bandara2022transformer} & 0.6214 & 0.4508 & 0.7182 & 0.5603 & 0.8685 & 0.7675 & 0.8232 & 0.6996 & 0.7593 & 0.6120  \\
DMINet \cite{feng2023change} & 0.5743 & 0.4028 & 0.6929 & 0.5302 & 0.8664 & 0.7642 & 0.8431 & 0.7288 & 0.7464 & 0.5953 \\
ICIFNet \cite{feng2022icif}& 0.5629 & 0.3917 & 0.6903 & 0.5270 & 0.8722 & 0.7734 & 0.8162 & 0.6895 & 0.7030 & 0.5421 \\
RDPNet \cite{chen2022rdp}& 0.5431 & 0.3728 & 0.6859 & 0.5219 & 0.8680 & 0.7668 & 0.8057 & 0.6747 & 0.7536 & 0.6046 \\
Siam-Diff \cite{daudt2018fully}& 0.4913 & 0.3257 & 0.6422 & 0.4729 & 0.8431 & 0.7288 & 0.7822 & 0.6423 & 0.5991 & 0.4277 \\
SNUNet \cite{fang2021snunet}& 0.5977 & 0.4262 & 0.7583 & 0.6106 & 0.8788 & 0.7838 & 0.8608 & 0.7556 & 0.7573 & 0.6073 \\ \midrule
Ours & \textbf{0.7939} & \textbf{0.6583} & \textbf{0.8553} & \textbf{0.7473} & \textbf{0.9023}  & \textbf{0.8220} & \textbf{0.9160} & \textbf{0.8450} & \textbf{0.8237} & \textbf{0.7003}  \\ \bottomrule
\end{tabular}
}
\label{CD_comp}
\end{table*}

The experiments discussed above demonstrate the superiority of our proposed method in multi-task learning settings. To further validate the performance of our model, we conduct a comparative analysis between ChangeMinds and other CD models. The experimental results are presented in Table \ref{CD_comp}. We can observe that our method outperforms all other methods across all five datasets. Specifically, on the GVLM and LEVIR-CD datasets, our method achieves improvements of 2.35\% and 2.76\% in F1 score, respectively, and improvements of 9.35\% and 4.58\% in cIoU, respectively, compared to the previous best methods. Notably, our method achieves significant improvements of 16.68\%, 9.35\%, and 8.83\% on the remaining CLCD, EGY-BCD, and SYSU-CD datasets in cIoU. These results indicate that our method can predict changes more accurately.

\begin{figure*}
		\centering
		\begin{minipage}{0.08\linewidth}
			\vspace{1pt}
			\centerline{\includegraphics[width=\textwidth]{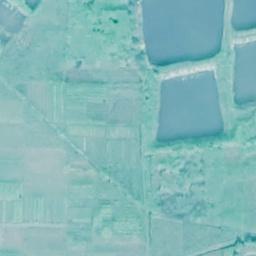}}
			\vspace{1pt}
			\centerline{\includegraphics[width=\textwidth]{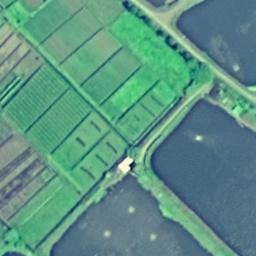}}
                \vspace{1pt}
		      \centerline{\includegraphics[width=\textwidth]{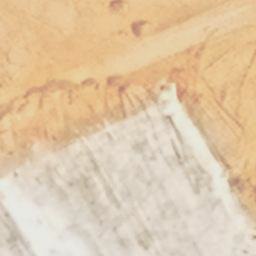}}
		      \vspace{1pt}
		      \centerline{\includegraphics[width=\textwidth]{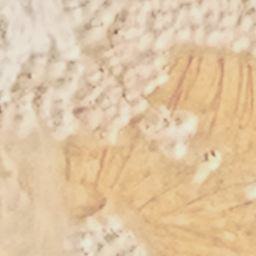}}
        		\centerline{\includegraphics[width=\textwidth]{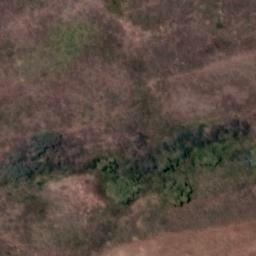}}
		\vspace{1pt}
		\centerline{\includegraphics[width=\textwidth]{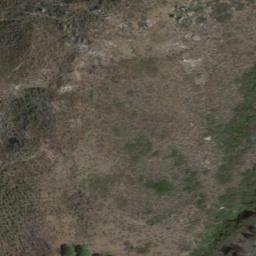}}
		\vspace{1pt}
		\centerline{\includegraphics[width=\textwidth]{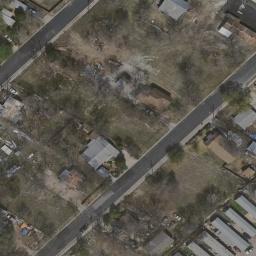}}
		\vspace{1pt}
		\centerline{\includegraphics[width=\textwidth]{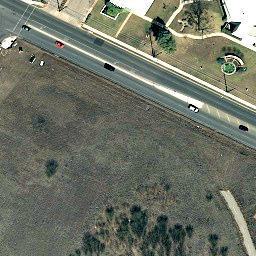}}
  		\vspace{1pt}
		\centerline{\includegraphics[width=\textwidth]{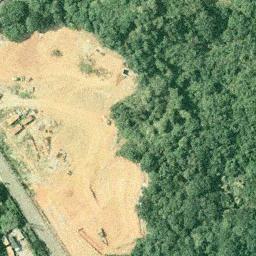}}
		\vspace{1pt}
		\centerline{\includegraphics[width=\textwidth]{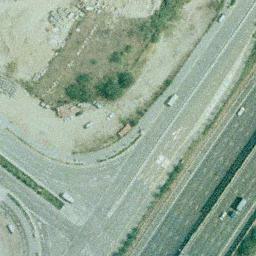}}
                \vspace{1pt}
			\centerline{(a)}
		\end{minipage}
		\hspace{-5pt}
		\begin{minipage}{0.08\linewidth}
			\vspace{1pt}
			\centerline{\includegraphics[width=\textwidth]{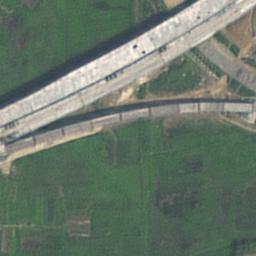}}
			\vspace{1pt}
			\centerline{\includegraphics[width=\textwidth]{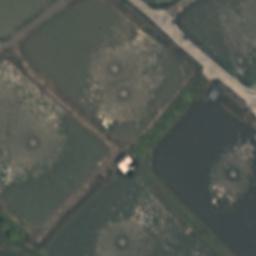}}
                \vspace{1pt}
		      \centerline{\includegraphics[width=\textwidth]{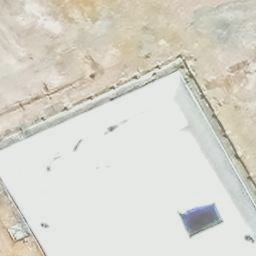}}
		      \vspace{1pt}
		      \centerline{\includegraphics[width=\textwidth]{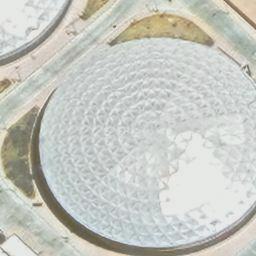}}
        		\centerline{\includegraphics[width=\textwidth]{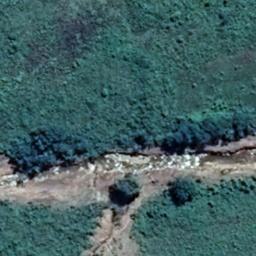}}
		\vspace{1pt}
		\centerline{\includegraphics[width=\textwidth]{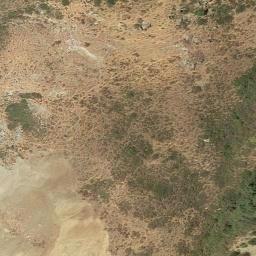}}
  		\vspace{1pt}
		\centerline{\includegraphics[width=\textwidth]{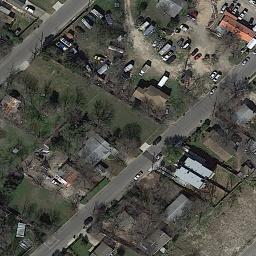}}
		\vspace{1pt}
		\centerline{\includegraphics[width=\textwidth]{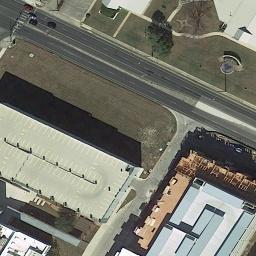}}
  		\vspace{1pt}
		\centerline{\includegraphics[width=\textwidth]{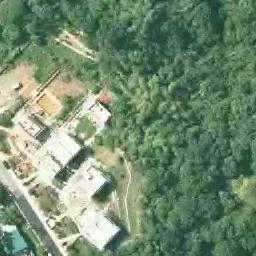}}
		\vspace{1pt}
		\centerline{\includegraphics[width=\textwidth]{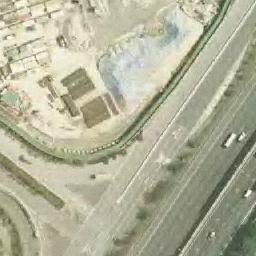}}
			\vspace{1pt}
			\centerline{(b)}
		\end{minipage}
		\hspace{-5pt}
		\begin{minipage}{0.08\linewidth}
			\vspace{1pt}
			\centerline{\includegraphics[width=\textwidth]{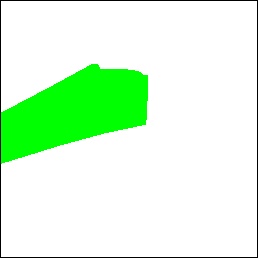}}
			\vspace{1pt}
			\centerline{\includegraphics[width=\textwidth]{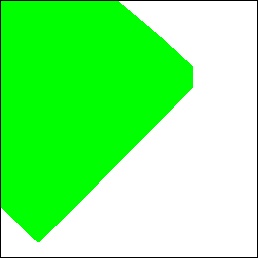}}
                \vspace{1pt}
		      \centerline{\includegraphics[width=\textwidth]{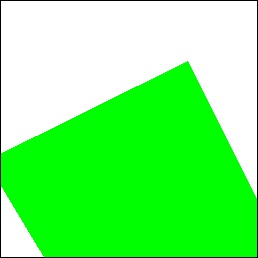}}
		      \vspace{1pt}
		      \centerline{\includegraphics[width=\textwidth]{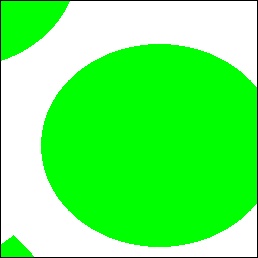}}
        		\centerline{\includegraphics[width=\textwidth]{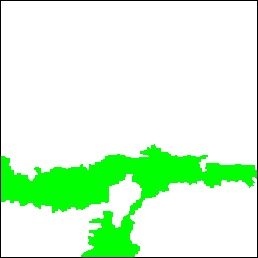}}
		\vspace{1pt}
		\centerline{\includegraphics[width=\textwidth]{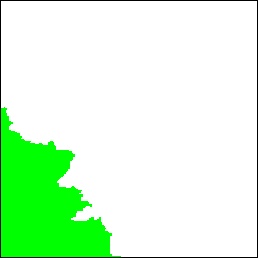}}
		\vspace{1pt}
		\centerline{\includegraphics[width=\textwidth]{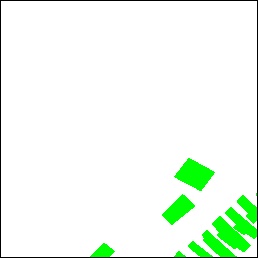}}
		\vspace{1pt}
		\centerline{\includegraphics[width=\textwidth]{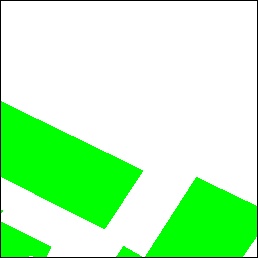}}
  		\vspace{1pt}
		\centerline{\includegraphics[width=\textwidth]{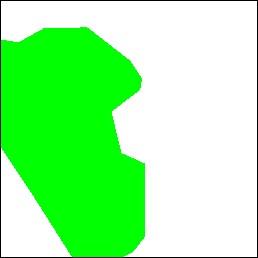}}
		\vspace{1pt}
		\centerline{\includegraphics[width=\textwidth]{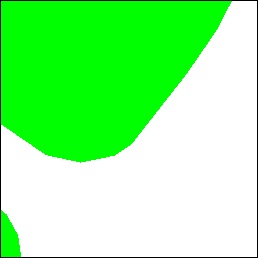}}
			\vspace{1pt}
			\centerline{(c)}
		\end{minipage}
		\hspace{-5pt}
		\begin{minipage}{0.08\linewidth}
			\vspace{1pt}
			\centerline{\includegraphics[width=\textwidth]{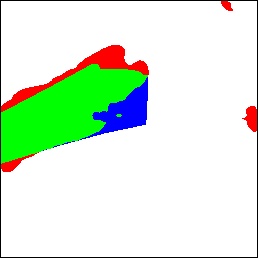}}
			\vspace{1pt}
			\centerline{\includegraphics[width=\textwidth]{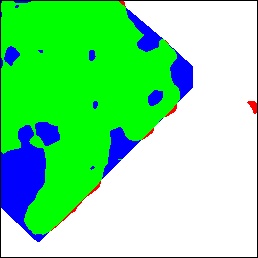}}
                \vspace{1pt}
		      \centerline{\includegraphics[width=\textwidth]{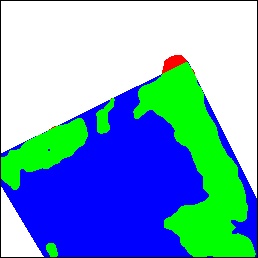}}
		      \vspace{1pt}
		      \centerline{\includegraphics[width=\textwidth]{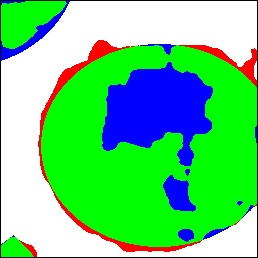}}
        		\centerline{\includegraphics[width=\textwidth]{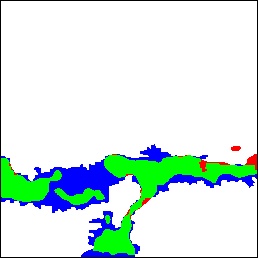}}
		\vspace{1pt}
		\centerline{\includegraphics[width=\textwidth]{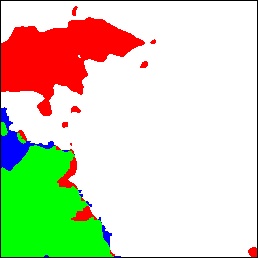}}
    		\vspace{1pt}
		\centerline{\includegraphics[width=\textwidth]{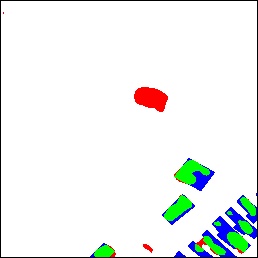}}
		\vspace{1pt}
		\centerline{\includegraphics[width=\textwidth]{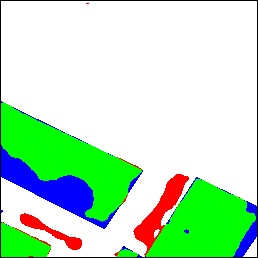}}
  		\vspace{1pt}
		\centerline{\includegraphics[width=\textwidth]{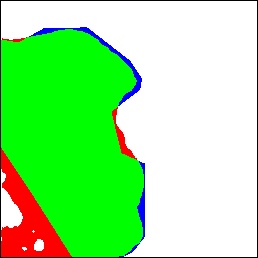}}
		\vspace{1pt}
		\centerline{\includegraphics[width=\textwidth]{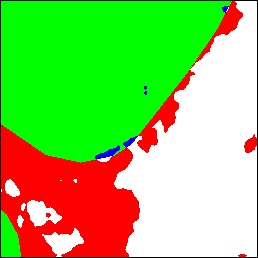}}
			\vspace{1pt}
			\centerline{(d)}
		\end{minipage}
		\hspace{-5pt}
		\begin{minipage}{0.08\linewidth}
			\vspace{1pt}
			\centerline{\includegraphics[width=\textwidth]{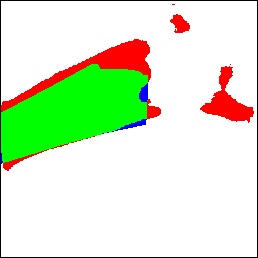}}
			\vspace{1pt}
			\centerline{\includegraphics[width=\textwidth]{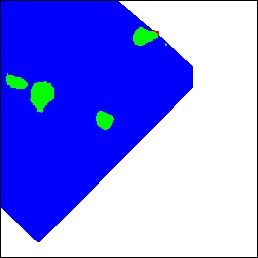}}
                \vspace{1pt}
		      \centerline{\includegraphics[width=\textwidth]{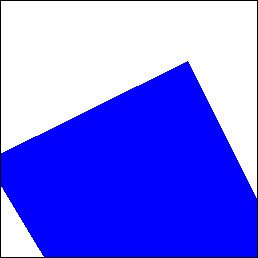}}
		      \vspace{1pt}
		      \centerline{\includegraphics[width=\textwidth]{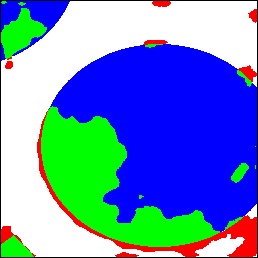}}
        		\centerline{\includegraphics[width=\textwidth]{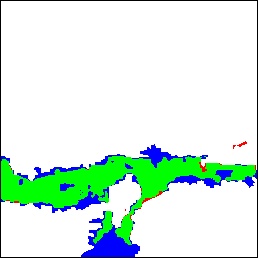}}
		\vspace{1pt}
		\centerline{\includegraphics[width=\textwidth]{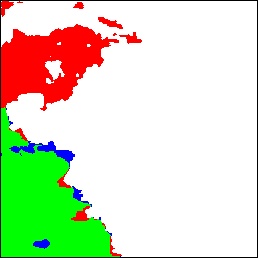}}
    		\vspace{1pt}
		\centerline{\includegraphics[width=\textwidth]{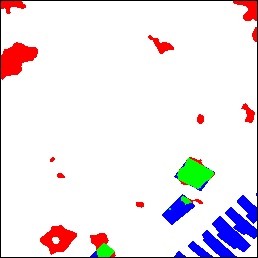}}
		\vspace{1pt}
		\centerline{\includegraphics[width=\textwidth]{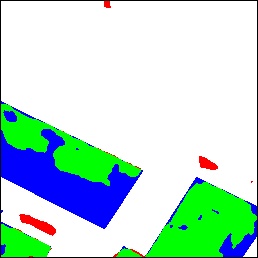}}
  		\vspace{1pt}
		\centerline{\includegraphics[width=\textwidth]{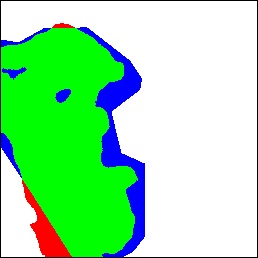}}
		\vspace{1pt}
		\centerline{\includegraphics[width=\textwidth]{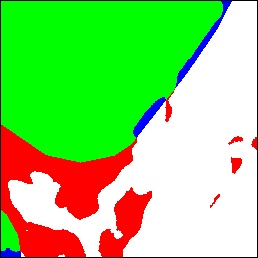}}
			\vspace{1pt}
			\centerline{(e)}
		\end{minipage}
  \hspace{-5pt}
		\begin{minipage}{0.08\linewidth}
			\vspace{1pt}
			\centerline{\includegraphics[width=\textwidth]{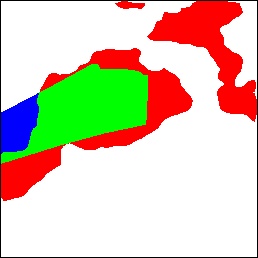}}
			\vspace{1pt}
			\centerline{\includegraphics[width=\textwidth]{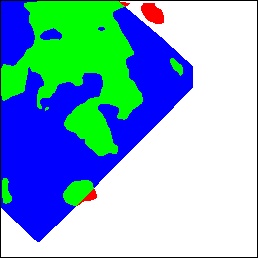}}
                \vspace{1pt}
		      \centerline{\includegraphics[width=\textwidth]{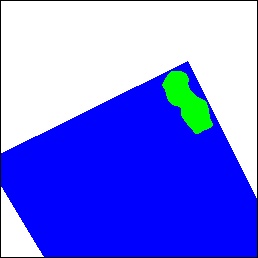}}
		      \vspace{1pt}
		      \centerline{\includegraphics[width=\textwidth]{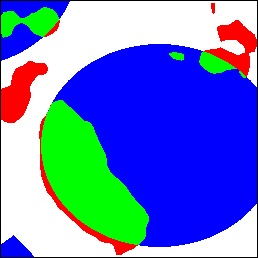}}
        		\centerline{\includegraphics[width=\textwidth]{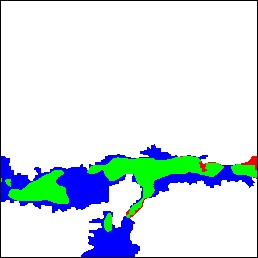}}
		\vspace{1pt}
		\centerline{\includegraphics[width=\textwidth]{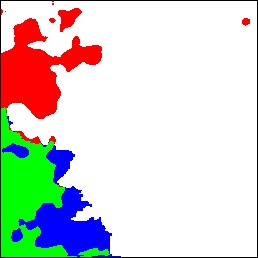}}
    		\vspace{1pt}
		\centerline{\includegraphics[width=\textwidth]{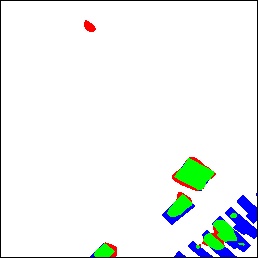}}
		\vspace{1pt}
		\centerline{\includegraphics[width=\textwidth]{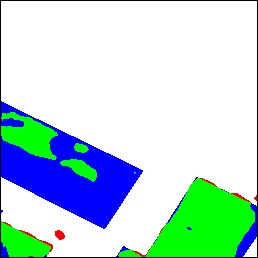}}
  		\vspace{1pt}
		\centerline{\includegraphics[width=\textwidth]{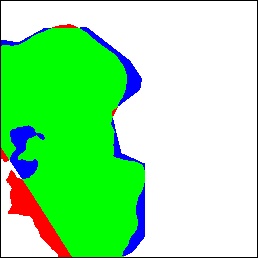}}
		\vspace{1pt}
		\centerline{\includegraphics[width=\textwidth]{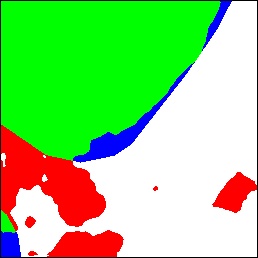}}
			\vspace{1pt}
			\centerline{(f)}
		\end{minipage}
    \hspace{-5pt}
	\begin{minipage}{0.08\linewidth}
		\vspace{1pt}
		\centerline{\includegraphics[width=\textwidth]{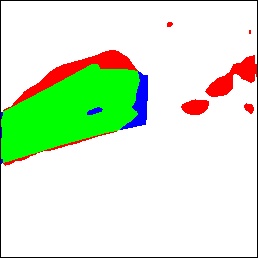}}
		\vspace{1pt}
		\centerline{\includegraphics[width=\textwidth]{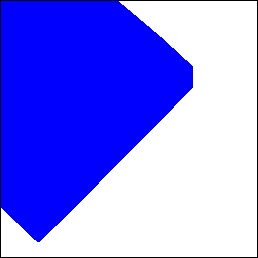}}
            \vspace{1pt}
		      \centerline{\includegraphics[width=\textwidth]{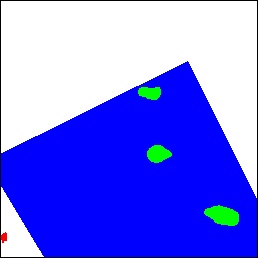}}
		      \vspace{1pt}
		      \centerline{\includegraphics[width=\textwidth]{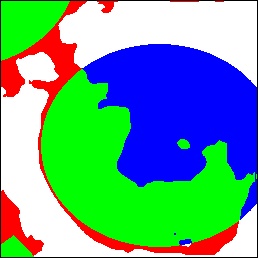}}
        		\centerline{\includegraphics[width=\textwidth]{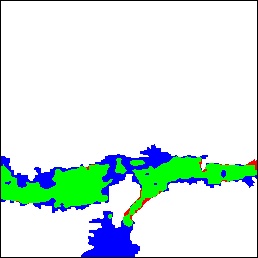}}
		\vspace{1pt}
		\centerline{\includegraphics[width=\textwidth]{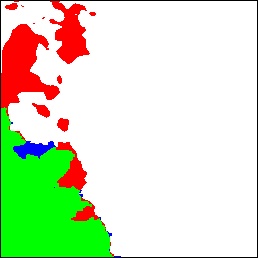}}
    		\vspace{1pt}
		\centerline{\includegraphics[width=\textwidth]{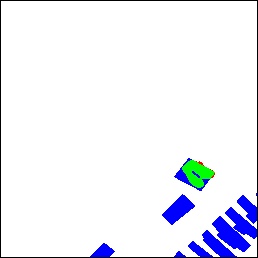}}
		\vspace{1pt}
		\centerline{\includegraphics[width=\textwidth]{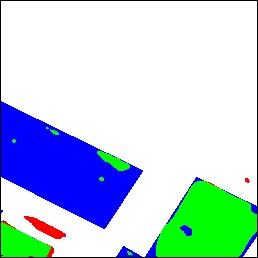}}
  		\vspace{1pt}
		\centerline{\includegraphics[width=\textwidth]{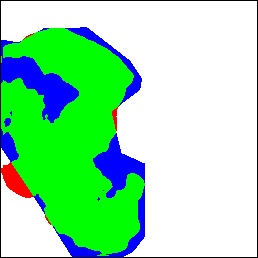}}
		\vspace{1pt}
		\centerline{\includegraphics[width=\textwidth]{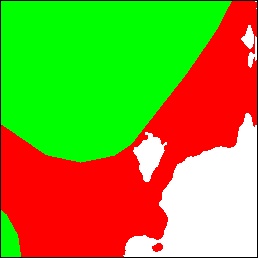}}
		\vspace{1pt}
		\centerline{(g)}
	\end{minipage}
  \hspace{-5pt}
  \begin{minipage}{0.08\linewidth}
	\vspace{1pt}
	\centerline{\includegraphics[width=\textwidth]{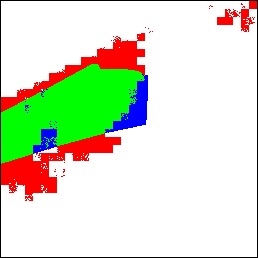}}
	\vspace{1pt}
	\centerline{\includegraphics[width=\textwidth]{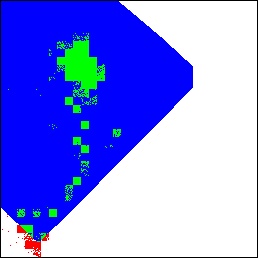}}
 \vspace{1pt}
		      \centerline{\includegraphics[width=\textwidth]{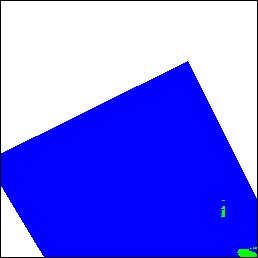}}
		      \vspace{1pt}
		      \centerline{\includegraphics[width=\textwidth]{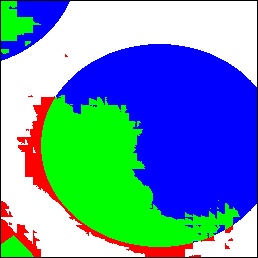}}
        		\centerline{\includegraphics[width=\textwidth]{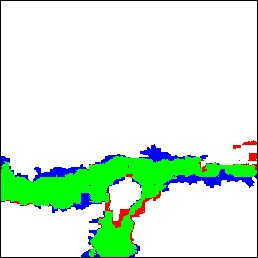}}
		\vspace{1pt}
		\centerline{\includegraphics[width=\textwidth]{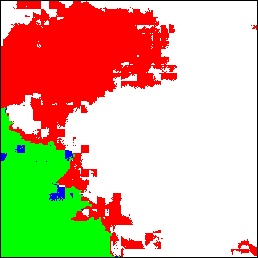}}
    		\vspace{1pt}
		\centerline{\includegraphics[width=\textwidth]{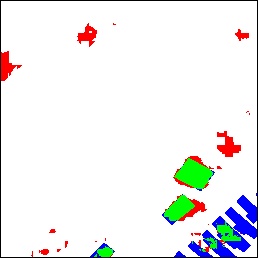}}
		\vspace{1pt}
		\centerline{\includegraphics[width=\textwidth]{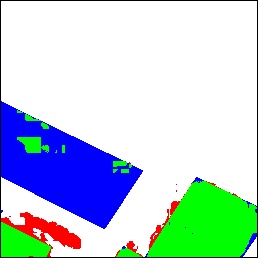}}
  		\vspace{1pt}
		\centerline{\includegraphics[width=\textwidth]{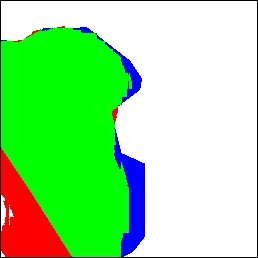}}
		\vspace{1pt}
		\centerline{\includegraphics[width=\textwidth]{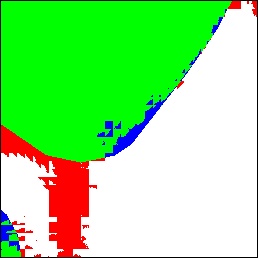}}
	\vspace{1pt}
	\centerline{(h)}
\end{minipage}
  \hspace{-5pt}
  \begin{minipage}{0.08\linewidth}
	\vspace{1pt}
	\centerline{\includegraphics[width=\textwidth]{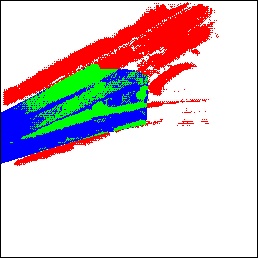}}
	\vspace{1pt}
	\centerline{\includegraphics[width=\textwidth]{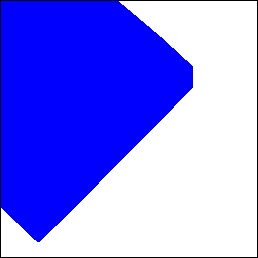}}
 \vspace{1pt}
		      \centerline{\includegraphics[width=\textwidth]{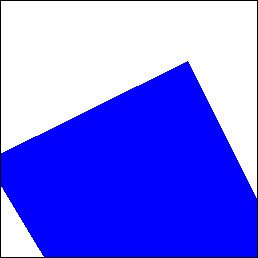}}
		      \vspace{1pt}
		      \centerline{\includegraphics[width=\textwidth]{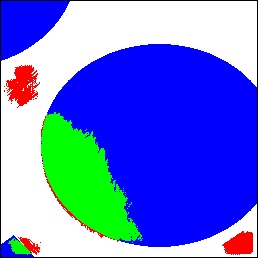}}
        		\centerline{\includegraphics[width=\textwidth]{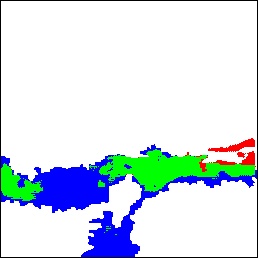}}
		\vspace{1pt}
		\centerline{\includegraphics[width=\textwidth]{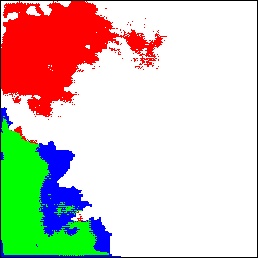}}
    		\vspace{1pt}
		\centerline{\includegraphics[width=\textwidth]{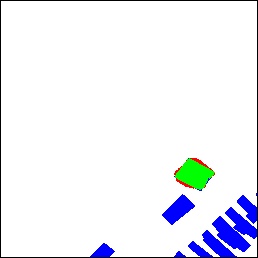}}
		\vspace{1pt}
		\centerline{\includegraphics[width=\textwidth]{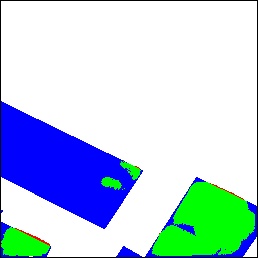}}
  		\vspace{1pt}
		\centerline{\includegraphics[width=\textwidth]{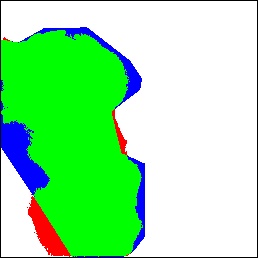}}
		\vspace{1pt}
		\centerline{\includegraphics[width=\textwidth]{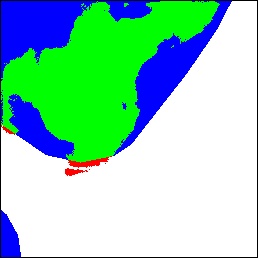}}
	\vspace{1pt}
	\centerline{(i)}
\end{minipage}
  \hspace{-5pt}
  \begin{minipage}{0.08\linewidth}
	\vspace{1pt}
	\centerline{\includegraphics[width=\textwidth]{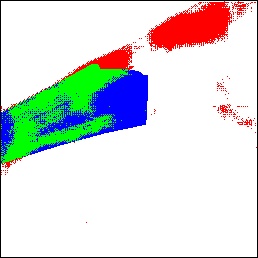}}
	\vspace{1pt}
	\centerline{\includegraphics[width=\textwidth]{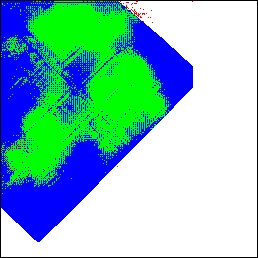}}
 \vspace{1pt}
		      \centerline{\includegraphics[width=\textwidth]{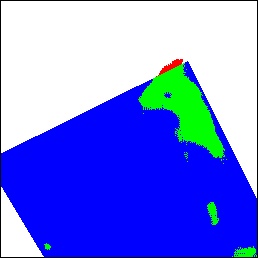}}
		      \vspace{1pt}
		      \centerline{\includegraphics[width=\textwidth]{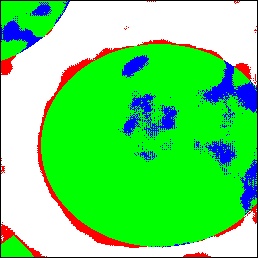}}
        		\centerline{\includegraphics[width=\textwidth]{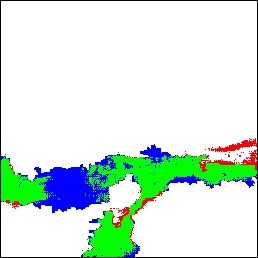}}
		\vspace{1pt}
		\centerline{\includegraphics[width=\textwidth]{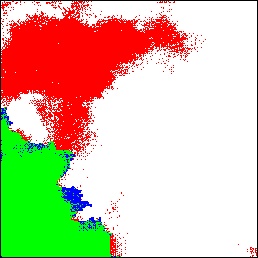}}
    		\vspace{1pt}
		\centerline{\includegraphics[width=\textwidth]{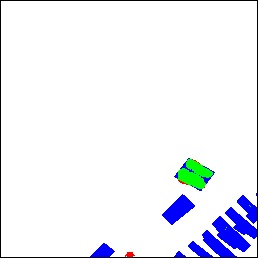}}
		\vspace{1pt}
		\centerline{\includegraphics[width=\textwidth]{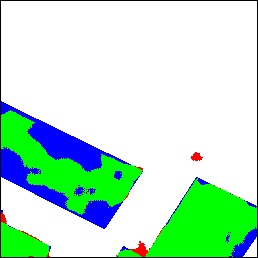}}
  		\vspace{1pt}
		\centerline{\includegraphics[width=\textwidth]{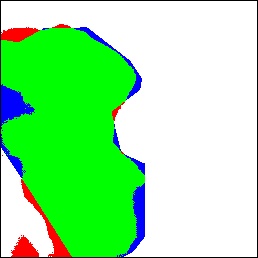}}
		\vspace{1pt}
		\centerline{\includegraphics[width=\textwidth]{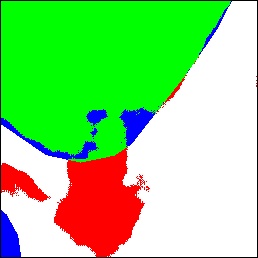}}
	\vspace{1pt}
	\centerline{(j)}
\end{minipage}
  \hspace{-5pt}
  \begin{minipage}{0.08\linewidth}
	\vspace{1pt}
	\centerline{\includegraphics[width=\textwidth]{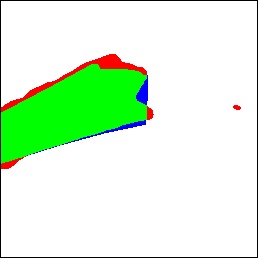}}
	\vspace{1pt}
	\centerline{\includegraphics[width=\textwidth]{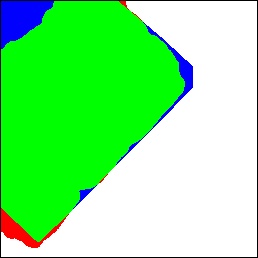}}
 \vspace{1pt}
		      \centerline{\includegraphics[width=\textwidth]{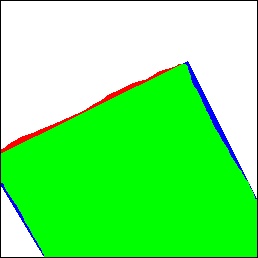}}
		      \vspace{1pt}
		      \centerline{\includegraphics[width=\textwidth]{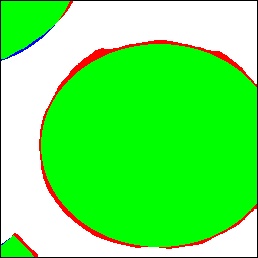}}
        		\centerline{\includegraphics[width=\textwidth]{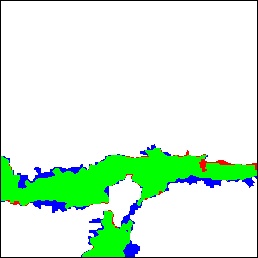}}
		\vspace{1pt}
		\centerline{\includegraphics[width=\textwidth]{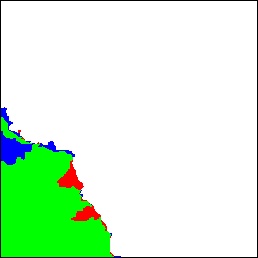}}
    		\vspace{1pt}
		\centerline{\includegraphics[width=\textwidth]{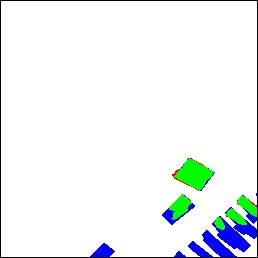}}
		\vspace{1pt}
		\centerline{\includegraphics[width=\textwidth]{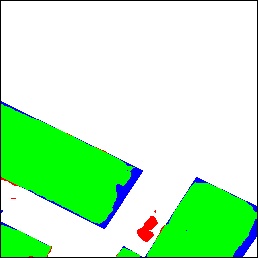}}
  		\vspace{1pt}
		\centerline{\includegraphics[width=\textwidth]{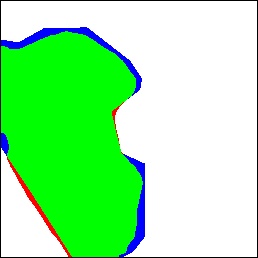}}
		\vspace{1pt}
		\centerline{\includegraphics[width=\textwidth]{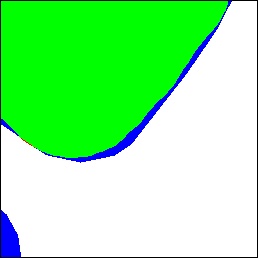}}
	\vspace{1pt}
	\centerline{(k)}
\end{minipage}
		\caption{Visual comparisons of the proposed method and the state-of-the-art approaches on five common CD datasets. (a) Pre-change images. (b) Post-change images. (c) Ground truth. (d) BIT. (e) ChangeFormer. (f) DMINet. (g) ICIFNet. (h) RDPnet. (i) SiamUnet-diff. (j) SNUNet. (k) ChangeMinds. The rendered colors represent true positives (green), true negatives (white), false positives (red), and false negatives (blue).} 
		\label{CD}
	\end{figure*}

Additionally, in Fig. \ref{CD}, we visualize the predicted change map comparison of ChangeMinds and other common CD methods across five different datasets, with two samples selected from each dataset. We can observe that, compared to other methods, ChangeMinds can more accurately capture subtle changes in small targets and generate much smoother change maps. These advantages demonstrate that our method can also achieve competitive performance in the single CD task.

\subsubsection{Comparison with other CC methods}

\begin{table}
\centering
\caption{Quantitative Comparison Results in terms of BLEU-1, BLEU-4, METEOR, ROUGE-L and CIDEr-D on DUBAI-CC Dataset. The best results are highlighted in bold.}
\label{CC_comp}
\resizebox{.5\linewidth}{!}{
\begin{tabular}{l|ccccc}
\toprule
Method & B1 & B4 & M & R &   C \\ 
\midrule
DUDA \cite{park2019robust} &0.5882 &0.2539 &0.2205 &0.4834 &0.6278\\
MCCFormer-S \cite{qiu2021describing} &0.5297 &0.2257 &0.1864 &0.4329 &0.5381 \\
MCCFormer-D \cite{qiu2021describing} &0.6465 &0.2948 &0.2509 &0.5127 &0.6651 \\
RSICCFormer \cite{liu2022remote} &0.6792 &0.3128 &0.2541 &0.5196 &0.6654 \\
Chg2Cap \cite{chang2023changes} & 0.6836  &0.3770  & 0.2783 & 0.5684   & 0.8474 \\ 
\midrule
Ours & \textbf{0.7305}	&\textbf{0.4008} &\textbf{0.2830}	 &\textbf{0.5742}	& \textbf{0.9791}	\\
\bottomrule
\end{tabular}
}
\end{table}

\begin{figure}
    \centering
    \includegraphics[width=.5\linewidth]{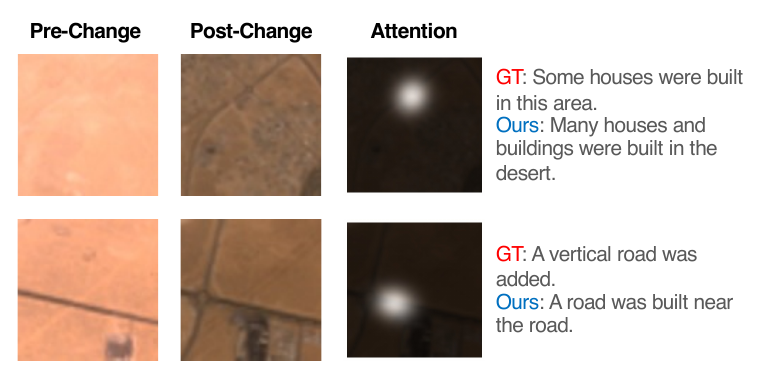}
    \caption{Visualized attention weights corresponding to the specific generated word in the CC classifier on the DUBAI-CC dataset. The corresponding words are marked in green.}
    \label{fig_DUBAI}
\end{figure}

In this section, we validate the performance of our proposed method in the single CC task using the DUBAI-CC dataset. The result is illustrated in Table \ref{CC_comp}. Our method outperforms previous models across all evaluation metrics. Specifically, for the two most critical CC task metrics, BLEU-4 and CIDEr-D, our method achieves improvements of 2.38\% and 13.17\% over Chg2Cap, respectively. This indicates that our approach can generate more precise change captions, suggesting that fusing multi-level deep features can improve the generation ability of the CC classifier.

Additionally, we visualize the captions generated by our proposed model on the DUBAI-CC dataset along with the corresponding attention weights in Fig. \ref{fig_DUBAI}. Our method can accurately pinpoint the regions where the small targets (houses and roads) are located. Additionally, the model successfully described the background of bi-temporal images in the caption generated in the first row and correctly identified the number of roads as two in the second row.

\section{Conclusion}\label{s5}

In this paper, we propose an end-to-end multi-task learning framework, \textit{ChangeMinds}, for detecting and describing changes in RS. Our ChangeMinds employs a Transformer-based Siamese encoder to extract deep features from bi-temporal images. Furthermore, we introduce the ChangeLSTM module, which can accurately model the intricate spatiotemporal patterns across bi-temporal deep features from two different directions. Meanwhile, we propose the Multi-task Predictor that uses the unified change decoder to fuse change-aware representations and employs multi-task classifiers to generate predictions for both CD and CC tasks.
The unified change decoder realizes multi-level feature aggregation and multi-task classifiers use a cross-attention layer to associate image features with text tokens. Extensive experiments on the LEVIR-MCI dataset and other common CC and CD datasets demonstrate that ChangeMinds outperforms previous state-of-the-art methods.

\bibliographystyle{IEEEtranN}
\bibliography{Ref}

\begin{thebibliography}{50}
\providecommand{\natexlab}[1]{#1}
\providecommand{\url}[1]{#1}
\csname url@samestyle\endcsname
\providecommand{\newblock}{\relax}
\providecommand{\bibinfo}[2]{#2}
\providecommand{\BIBentrySTDinterwordspacing}{\spaceskip=0pt\relax}
\providecommand{\BIBentryALTinterwordstretchfactor}{4}
\providecommand{\BIBentryALTinterwordspacing}{\spaceskip=\fontdimen2\font plus
\BIBentryALTinterwordstretchfactor\fontdimen3\font minus \fontdimen4\font\relax}
\providecommand{\BIBforeignlanguage}[2]{{%
\expandafter\ifx\csname l@#1\endcsname\relax
\typeout{** WARNING: IEEEtranN.bst: No hyphenation pattern has been}%
\typeout{** loaded for the language `#1'. Using the pattern for}%
\typeout{** the default language instead.}%
\else
\language=\csname l@#1\endcsname
\fi
#2}}
\providecommand{\BIBdecl}{\relax}
\BIBdecl

\bibitem[Ghamisi et~al.(2019)Ghamisi, Rasti, Yokoya, Wang, Hofle, Bruzzone, Bovolo, Chi, Anders, Gloaguen, Atkinson, and Benediktsson]{8672156}
P.~Ghamisi, B.~Rasti, N.~Yokoya, Q.~Wang, B.~Hofle, L.~Bruzzone, F.~Bovolo, M.~Chi, K.~Anders, R.~Gloaguen, P.~M. Atkinson, and J.~A. Benediktsson, ``Multisource and multitemporal data fusion in remote sensing: A comprehensive review of the state of the art,'' \emph{IEEE Geoscience and Remote Sensing Magazine}, vol.~7, no.~1, pp. 6--39, 2019.

\bibitem[Ghamisi et~al.(2024)Ghamisi, Yu, Marinoni, Gevaert, Persello, Selvakumaran, Girotto, Horton, Rufin, Hostert, et~al.]{ghamisi2024responsible}
P.~Ghamisi, W.~Yu, A.~Marinoni, C.~M. Gevaert, C.~Persello, S.~Selvakumaran, M.~Girotto, B.~P. Horton, P.~Rufin, P.~Hostert \emph{et~al.}, ``Responsible ai for earth observation,'' \emph{arXiv preprint arXiv:2405.20868}, 2024.

\bibitem[Li et~al.(2022)Li, Lu, Zhang, Tu, Li, Huang, Robinson, Malkin, Jojic, Ghamisi, Hänsch, and Yokoya]{9690575}
Z.~Li, F.~Lu, H.~Zhang, L.~Tu, J.~Li, X.~Huang, C.~Robinson, N.~Malkin, N.~Jojic, P.~Ghamisi, R.~Hänsch, and N.~Yokoya, ``The outcome of the 2021 ieee grss data fusion contest—track msd: Multitemporal semantic change detection,'' \emph{IEEE Journal of Selected Topics in Applied Earth Observations and Remote Sensing}, vol.~15, pp. 1643--1655, 2022.

\bibitem[Zhang et~al.(2023)Zhang, Yu, Pun, and Shi]{zhang2023cross}
X.~Zhang, W.~Yu, M.-O. Pun, and W.~Shi, ``Cross-domain landslide mapping from large-scale remote sensing images using prototype-guided domain-aware progressive representation learning,'' \emph{ISPRS Journal of Photogrammetry and Remote Sensing}, vol. 197, pp. 1--17, 2023.

\bibitem[Yu et~al.(2024{\natexlab{a}})Yu, Zhang, Das, Xiang~Zhu, and Ghamisi]{yu2024maskcd}
W.~Yu, X.~Zhang, S.~Das, X.~Xiang~Zhu, and P.~Ghamisi, ``Maskcd: A remote sensing change detection network based on mask classification,'' \emph{IEEE Transactions on Geoscience and Remote Sensing}, vol.~62, pp. 1--16, 2024.

\bibitem[Chang et~al.(2024)Chang, Wang, Diao, Xu, and Sun]{chang2024remote}
H.~Chang, P.~Wang, W.~Diao, G.~Xu, and X.~Sun, ``Remote sensing change detection with bitemporal and differential feature interactive perception,'' \emph{IEEE Transactions on Image Processing}, 2024.

\bibitem[Hu et~al.(2023)Hu, Wu, Du, and Zhang]{hu2023binary}
M.~Hu, C.~Wu, B.~Du, and L.~Zhang, ``Binary change guided hyperspectral multiclass change detection,'' \emph{IEEE Transactions on Image Processing}, vol.~32, pp. 791--806, 2023.

\bibitem[Lei et~al.(2020)Lei, Peng, Zhang, Ke, and Li]{lei2020hierarchical}
Y.~Lei, D.~Peng, P.~Zhang, Q.~Ke, and H.~Li, ``Hierarchical paired channel fusion network for street scene change detection,'' \emph{IEEE Transactions on Image Processing}, vol.~30, pp. 55--67, 2020.

\bibitem[Liu et~al.(2017)Liu, Li, Mercier, He, and Pan]{liu2017change}
Z.~Liu, G.~Li, G.~Mercier, Y.~He, and Q.~Pan, ``Change detection in heterogenous remote sensing images via homogeneous pixel transformation,'' \emph{IEEE Transactions on Image Processing}, vol.~27, no.~4, pp. 1822--1834, 2017.

\bibitem[Tu et~al.(2023)Tu, Li, Su, Du, Lu, and Huang]{tu2023adaptive}
Y.~Tu, L.~Li, L.~Su, J.~Du, K.~Lu, and Q.~Huang, ``Adaptive representation disentanglement network for change captioning,'' \emph{IEEE Transactions on Image Processing}, vol.~32, pp. 2620--2635, 2023.

\bibitem[Liu et~al.(2024)Liu, Chen, Zhang, Qi, Zou, and Shi]{liu2024change}
C.~Liu, K.~Chen, H.~Zhang, Z.~Qi, Z.~Zou, and Z.~Shi, ``Change-agent: Towards interactive comprehensive remote sensing change interpretation and analysis,'' \emph{IEEE Transactions on Geoscience and Remote Sensing}, 2024.

\bibitem[Wu et~al.(2024)Wu, Zhang, Du, Chen, Wang, and Zhong]{10616141}
C.~Wu, L.~Zhang, B.~Du, H.~Chen, J.~Wang, and H.~Zhong, ``Unet-like remote sensing change detection: A review of current models and research directions,'' \emph{IEEE Geoscience and Remote Sensing Magazine}, pp. 2--31, 2024.

\bibitem[Daudt et~al.(2018)Daudt, Le~Saux, and Boulch]{daudt2018fully}
R.~C. Daudt, B.~Le~Saux, and A.~Boulch, ``Fully convolutional siamese networks for change detection,'' in \emph{2018 25th IEEE international conference on image processing (ICIP)}.\hskip 1em plus 0.5em minus 0.4em\relax IEEE, 2018, pp. 4063--4067.

\bibitem[Zhang et~al.(2022)Zhang, Yu, and Pun]{zhang2022multilevel}
X.~Zhang, W.~Yu, and M.-O. Pun, ``Multilevel deformable attention-aggregated networks for change detection in bitemporal remote sensing imagery,'' \emph{IEEE Transactions on Geoscience and Remote Sensing}, vol.~60, pp. 1--18, 2022.

\bibitem[Chen et~al.(2024)Chen, Song, Han, Xia, and Yokoya]{10565926}
H.~Chen, J.~Song, C.~Han, J.~Xia, and N.~Yokoya, ``Changemamba: Remote sensing change detection with spatiotemporal state space model,'' \emph{IEEE Transactions on Geoscience and Remote Sensing}, vol.~62, pp. 1--20, 2024.

\bibitem[Yu et~al.(2024{\natexlab{b}})Yu, Zhang, Zhu, Gloaguen, and Ghamisi]{yu2024minenetcd}
W.~Yu, X.~Zhang, X.~X. Zhu, R.~Gloaguen, and P.~Ghamisi, ``Minenetcd: A benchmark for global mining change detection on remote sensing imagery,'' \emph{arXiv preprint arXiv:2407.03971}, 2024.

\bibitem[Beck et~al.(2024)Beck, P{\"o}ppel, Spanring, Auer, Prudnikova, Kopp, Klambauer, Brandstetter, and Hochreiter]{beck2024xlstm}
M.~Beck, K.~P{\"o}ppel, M.~Spanring, A.~Auer, O.~Prudnikova, M.~Kopp, G.~Klambauer, J.~Brandstetter, and S.~Hochreiter, ``xlstm: Extended long short-term memory,'' \emph{arXiv preprint arXiv:2405.04517}, 2024.

\bibitem[Chouaf et~al.(2021)Chouaf, Hoxha, Smara, and Melgani]{chouaf2021captioning}
S.~Chouaf, G.~Hoxha, Y.~Smara, and F.~Melgani, ``Captioning changes in bi-temporal remote sensing images,'' in \emph{2021 IEEE International Geoscience and Remote Sensing Symposium IGARSS}.\hskip 1em plus 0.5em minus 0.4em\relax IEEE, 2021, pp. 2891--2894.

\bibitem[Liu et~al.(2022{\natexlab{a}})Liu, Zhao, Chen, Zou, and Shi]{liu2022remote}
C.~Liu, R.~Zhao, H.~Chen, Z.~Zou, and Z.~Shi, ``Remote sensing image change captioning with dual-branch transformers: A new method and a large scale dataset,'' \emph{IEEE Transactions on Geoscience and Remote Sensing}, vol.~60, pp. 1--20, 2022.

\bibitem[Chen and Shi(2020)]{chen2020spatial}
H.~Chen and Z.~Shi, ``A spatial-temporal attention-based method and a new dataset for remote sensing image change detection,'' \emph{Remote Sensing}, vol.~12, no.~10, p. 1662, 2020.

\bibitem[Hoxha et~al.(2022)Hoxha, Chouaf, Melgani, and Smara]{hoxha2022change}
G.~Hoxha, S.~Chouaf, F.~Melgani, and Y.~Smara, ``Change captioning: A new paradigm for multitemporal remote sensing image analysis,'' \emph{IEEE Transactions on Geoscience and Remote Sensing}, vol.~60, pp. 1--14, 2022.

\bibitem[Chang and Ghamisi(2023)]{chang2023changes}
S.~Chang and P.~Ghamisi, ``Changes to captions: An attentive network for remote sensing change captioning,'' \emph{IEEE Transactions on Image Processing}, 2023.

\bibitem[Liu et~al.(2023{\natexlab{a}})Liu, Zhao, Chen, Qi, Zou, and Shi]{liu2023decoupling}
C.~Liu, R.~Zhao, J.~Chen, Z.~Qi, Z.~Zou, and Z.~Shi, ``A decoupling paradigm with prompt learning for remote sensing image change captioning,'' \emph{IEEE Transactions on Geoscience and Remote Sensing}, 2023.

\bibitem[Zhao et~al.(2021)Zhao, Shi, and Zou]{zhao2021high}
R.~Zhao, Z.~Shi, and Z.~Zou, ``High-resolution remote sensing image captioning based on structured attention,'' \emph{IEEE Transactions on Geoscience and Remote Sensing}, vol.~60, pp. 1--14, 2021.

\bibitem[Xu et~al.(2023)Xu, Yu, Ghamisi, Kopp, and Hochreiter]{xu2023txt2img}
Y.~Xu, W.~Yu, P.~Ghamisi, M.~Kopp, and S.~Hochreiter, ``Txt2img-mhn: Remote sensing image generation from text using modern hopfield networks,'' \emph{IEEE Transactions on Image Processing}, 2023.

\bibitem[Wang and Ghamisi(2024)]{wang2024rsadapter}
Y.~Wang and P.~Ghamisi, ``Rsadapter: Adapting multimodal models for remote sensing visual question answering,'' \emph{IEEE Transactions on Geoscience and Remote Sensing}, 2024.

\bibitem[Liu et~al.(2023{\natexlab{b}})Liu, Yang, Qi, Zou, and Shi]{liu2023progressive}
C.~Liu, J.~Yang, Z.~Qi, Z.~Zou, and Z.~Shi, ``Progressive scale-aware network for remote sensing image change captioning,'' in \emph{IGARSS 2023-2023 IEEE International Geoscience and Remote Sensing Symposium}.\hskip 1em plus 0.5em minus 0.4em\relax IEEE, 2023, pp. 6668--6671.

\bibitem[Liu et~al.(2021)Liu, Lin, Cao, Hu, Wei, Zhang, Lin, and Guo]{liu2021swin}
Z.~Liu, Y.~Lin, Y.~Cao, H.~Hu, Y.~Wei, Z.~Zhang, S.~Lin, and B.~Guo, ``Swin transformer: Hierarchical vision transformer using shifted windows,'' in \emph{Proceedings of the IEEE/CVF international conference on computer vision}, 2021, pp. 10\,012--10\,022.

\bibitem[Dosovitskiy et~al.(2020)Dosovitskiy, Beyer, Kolesnikov, Weissenborn, Zhai, Unterthiner, Dehghani, Minderer, Heigold, Gelly, et~al.]{dosovitskiy2020image}
A.~Dosovitskiy, L.~Beyer, A.~Kolesnikov, D.~Weissenborn, X.~Zhai, T.~Unterthiner, M.~Dehghani, M.~Minderer, G.~Heigold, S.~Gelly \emph{et~al.}, ``An image is worth 16x16 words: Transformers for image recognition at scale,'' in \emph{International Conference on Learning Representations}, 2020.

\bibitem[Alkin et~al.(2024)Alkin, Beck, P{\"o}ppel, Hochreiter, and Brandstetter]{alkin2024vision}
B.~Alkin, M.~Beck, K.~P{\"o}ppel, S.~Hochreiter, and J.~Brandstetter, ``Vision-lstm: xlstm as generic vision backbone,'' \emph{arXiv preprint arXiv:2406.04303}, 2024.

\bibitem[Elfwing et~al.(2018)Elfwing, Uchibe, and Doya]{elfwing2018sigmoid}
S.~Elfwing, E.~Uchibe, and K.~Doya, ``Sigmoid-weighted linear units for neural network function approximation in reinforcement learning,'' \emph{Neural networks}, vol. 107, pp. 3--11, 2018.

\bibitem[Xiao et~al.(2018)Xiao, Liu, Zhou, Jiang, and Sun]{xiao2018unified}
T.~Xiao, Y.~Liu, B.~Zhou, Y.~Jiang, and J.~Sun, ``Unified perceptual parsing for scene understanding,'' in \emph{Proceedings of the European conference on computer vision (ECCV)}, 2018, pp. 418--434.

\bibitem[He et~al.(2022)He, Feng, Cheng, Ji, Guo, and Caverlee]{he2022metabalance}
Y.~He, X.~Feng, C.~Cheng, G.~Ji, Y.~Guo, and J.~Caverlee, ``Metabalance: improving multi-task recommendations via adapting gradient magnitudes of auxiliary tasks,'' in \emph{Proceedings of the ACM Web Conference 2022}, 2022, pp. 2205--2215.

\bibitem[Liu et~al.(2022{\natexlab{b}})Liu, Chai, Deng, and Liu]{liu2022cnn}
M.~Liu, Z.~Chai, H.~Deng, and R.~Liu, ``A cnn-transformer network with multiscale context aggregation for fine-grained cropland change detection,'' \emph{IEEE Journal of Selected Topics in Applied Earth Observations and Remote Sensing}, vol.~15, pp. 4297--4306, 2022.

\bibitem[Holail et~al.(2023)Holail, Saleh, Xiao, and Li]{holail2023afde}
S.~Holail, T.~Saleh, X.~Xiao, and D.~Li, ``Afde-net: Building change detection using attention-based feature differential enhancement for satellite imagery,'' \emph{IEEE Geoscience and Remote Sensing Letters}, 2023.

\bibitem[Shi et~al.(2021)Shi, Liu, Li, Liu, Wang, and Zhang]{shi2021deeply}
Q.~Shi, M.~Liu, S.~Li, X.~Liu, F.~Wang, and L.~Zhang, ``A deeply supervised attention metric-based network and an open aerial image dataset for remote sensing change detection,'' \emph{IEEE transactions on geoscience and remote sensing}, vol.~60, pp. 1--16, 2021.

\bibitem[Chen et~al.(2021)Chen, Qi, and Shi]{chen2021remote}
H.~Chen, Z.~Qi, and Z.~Shi, ``Remote sensing image change detection with transformers,'' \emph{IEEE Transactions on Geoscience and Remote Sensing}, vol.~60, pp. 1--14, 2021.

\bibitem[Bandara and Patel(2022)]{bandara2022transformer}
W.~G.~C. Bandara and V.~M. Patel, ``A transformer-based siamese network for change detection,'' in \emph{IGARSS 2022-2022 IEEE International Geoscience and Remote Sensing Symposium}.\hskip 1em plus 0.5em minus 0.4em\relax IEEE, 2022, pp. 207--210.

\bibitem[Feng et~al.(2023)Feng, Jiang, Xu, and Zheng]{feng2023change}
Y.~Feng, J.~Jiang, H.~Xu, and J.~Zheng, ``Change detection on remote sensing images using dual-branch multilevel intertemporal network,'' \emph{IEEE Transactions on Geoscience and Remote Sensing}, vol.~61, pp. 1--15, 2023.

\bibitem[Feng et~al.(2022)Feng, Xu, Jiang, Liu, and Zheng]{feng2022icif}
Y.~Feng, H.~Xu, J.~Jiang, H.~Liu, and J.~Zheng, ``Icif-net: Intra-scale cross-interaction and inter-scale feature fusion network for bitemporal remote sensing images change detection,'' \emph{IEEE Transactions on Geoscience and Remote Sensing}, vol.~60, pp. 1--13, 2022.

\bibitem[Chen et~al.(2022)Chen, Pu, Yang, Tang, and Xu]{chen2022rdp}
H.~Chen, F.~Pu, R.~Yang, R.~Tang, and X.~Xu, ``Rdp-net: Region detail preserving network for change detection,'' \emph{IEEE Transactions on Geoscience and Remote Sensing}, vol.~60, pp. 1--10, 2022.

\bibitem[Fang et~al.(2021)Fang, Li, Shao, and Li]{fang2021snunet}
S.~Fang, K.~Li, J.~Shao, and Z.~Li, ``Snunet-cd: A densely connected siamese network for change detection of vhr images,'' \emph{IEEE Geoscience and Remote Sensing Letters}, vol.~19, pp. 1--5, 2021.

\bibitem[Park et~al.(2019)Park, Darrell, and Rohrbach]{park2019robust}
D.~H. Park, T.~Darrell, and A.~Rohrbach, ``Robust change captioning,'' in \emph{Proceedings of the IEEE/CVF International Conference on Computer Vision}, 2019, pp. 4624--4633.

\bibitem[Qiu et~al.(2021)Qiu, Yamamoto, Nakashima, Suzuki, Iwata, Kataoka, and Satoh]{qiu2021describing}
Y.~Qiu, S.~Yamamoto, K.~Nakashima, R.~Suzuki, K.~Iwata, H.~Kataoka, and Y.~Satoh, ``Describing and localizing multiple changes with transformers,'' in \emph{Proceedings of the IEEE/CVF International Conference on Computer Vision}, 2021, pp. 1971--1980.

\bibitem[Kingma(2014)]{kingma2014adam}
D.~P. Kingma, ``Adam: A method for stochastic optimization,'' \emph{arXiv preprint arXiv:1412.6980}, 2014.

\bibitem[Gugger et~al.(2022)Gugger, Debut, Wolf, Schmid, Mueller, Mangrulkar, Sun, and Bossan]{accelerate}
S.~Gugger, L.~Debut, T.~Wolf, P.~Schmid, Z.~Mueller, S.~Mangrulkar, M.~Sun, and B.~Bossan, ``Accelerate: Training and inference at scale made simple, efficient and adaptable.'' \url{https://github.com/huggingface/accelerate}, 2022.

\bibitem[Papineni et~al.(2002)Papineni, Roukos, Ward, and Zhu]{papineni2002bleu}
K.~Papineni, S.~Roukos, T.~Ward, and W.-J. Zhu, ``Bleu: a method for automatic evaluation of machine translation,'' in \emph{Proceedings of the 40th annual meeting of the Association for Computational Linguistics}, 2002, pp. 311--318.

\bibitem[Banerjee and Lavie(2005)]{banerjee2005meteor}
S.~Banerjee and A.~Lavie, ``Meteor: An automatic metric for mt evaluation with improved correlation with human judgments,'' in \emph{Proceedings of the acl workshop on intrinsic and extrinsic evaluation measures for machine translation and/or summarization}, 2005, pp. 65--72.

\bibitem[Lin(2004)]{lin2004rouge}
C.-Y. Lin, ``Rouge: A package for automatic evaluation of summaries,'' in \emph{Text summarization branches out}, 2004, pp. 74--81.

\bibitem[Vedantam et~al.(2015)Vedantam, Lawrence~Zitnick, and Parikh]{vedantam2015cider}
R.~Vedantam, C.~Lawrence~Zitnick, and D.~Parikh, ``Cider: Consensus-based image description evaluation,'' in \emph{Proceedings of the IEEE conference on computer vision and pattern recognition}, 2015, pp. 4566--4575.

\end{thebibliography}
\end{document}